\documentclass{article}

\usepackage[preprint]{colm}

\usepackage{microtype}
\usepackage{tabularx}
\usepackage{hyperref}
\usepackage{url}
\usepackage{booktabs}
\usepackage{graphicx}
\usepackage{lineno}
\usepackage{color}
\usepackage{amssymb}
\usepackage{paralist}

\usepackage{url}
\usepackage{breakurl}  % 允许URL换行
\usepackage{hyperref}

\usepackage{subcaption}
\usepackage{graphicx}
\usepackage{latexsym}
\usepackage{amsmath}
\usepackage{float}
\usepackage{footnote}
\usepackage{enumitem}
\usepackage{bm}
\usepackage{arydshln}
\usepackage{booktabs}
\usepackage{multicol}
\usepackage{multirow}
\usepackage{color} 
\usepackage{colortbl}
\usepackage{bbding}
\usepackage{makecell}
\usepackage{mathtools}
\usepackage{imakeidx}
\usepackage{longtable}
\usepackage{wrapfig}
\usepackage{rotating}

\usepackage{arydshln}
\usepackage{lipsum}
\usepackage{natbib}
\usepackage[toc]{multitoc}
\usepackage[edges]{forest}
\usepackage[normalem]{ulem}
\definecolor{mydarkblue}{rgb}{0,0.08,0.45}
\usepackage{CJKutf8}
\usepackage{awesomebox} % for infobox
\usepackage{bbding}
\usepackage[most]{tcolorbox}
\usepackage{booktabs}
\usepackage{pifont}
\newcommand{\cmark}{\ding{51}} % green with \textcolor if desired

\definecolor{wkblue}{rgb}{0.2, 0.3, 0.6}
\definecolor{meta-color}{rgb}{0.5, 0.5, 0.5}
\usepackage{amsmath}
\usepackage{enumitem}
\usepackage{lscape} 
\usepackage{booktabs}
\usepackage{algorithm}      % For the algorithm environment
\usepackage{algorithmicx}
\usepackage{algpseudocode}
\usepackage{tabularx,booktabs}
\usepackage{makecell}

\usepackage{fancyvrb}
\usepackage{fvextra}

\usepackage{amsfonts}

\definecolor{darkblue}{rgb}{0, 0, 0.5}
\hypersetup{
    colorlinks=true,
    citecolor=darkblue,
    linkcolor=darkblue,
    urlcolor=darkblue
}

\DeclareCaptionFont{black}{\color{black}}

\definecolor{geovistagray}{gray}{0.95}
\definecolor{myblue}{rgb}{0.9, 0.1, 0.94}
\definecolor{mygreen}{rgb}{0.64, 0.56, 0.88}
\definecolor{myyellow}{rgb}{0.68, 0.6, 0.1}
\definecolor{fancygreen}{rgb}{0.33, 0.68, 0.20}
\definecolor{salmon}{rgb}{0.94, 0.52, 0.49}
\definecolor{tablegreen}{rgb}{0.82, 0.94, 0.75}
\definecolor{tableblue}{rgb}{0.81, 0.90, 0.94}
\definecolor{tablered}{rgb}{0.97, 0.85, 0.85}
\definecolor{tableorange}{rgb}{0.96, 0.85, 0.81}

\newenvironment{itemize*}%
 {\leftmargini=10pt\begin{itemize}%
  \setlength{\itemsep}{0pt}%
  \setlength{\parskip}{0pt}%
  }%
 {\end{itemize}}
\newenvironment{enumerate*}%
 {\begin{enumerate}%
  \setlength{\itemsep}{0pt}%
  \setlength{\parskip}{0pt}}%
 {\end{enumerate}}

\title{GeoVista: Web-Augmented \\
Agentic Visual Reasoning for Geolocalization}

\author{Yikun Wang$^{1,4}$, Zuyan Liu$^{3}$, Ziyi Wang$^{3}$, Han Hu$^{2}$, Pengfei Liu$^{4}$, Yongming Rao$^{2}$\thanks{Corresponding author}
\\
$^1$Fudan University \quad $^2$Tencent Hunyuan \\
$^3$Tsinghua University \quad 
$^4$Shanghai Innovation Institute
}

\begin{document}
\maketitle

\maketitle

    \begin{abstract}

        Current research on agentic visual reasoning enables deep multimodal understanding but primarily focuses on image manipulation tools, leaving a gap toward more general-purpose agentic models. In this work, we revisit the geolocalization task, which requires not only nuanced visual grounding but also web search to confirm or refine hypotheses during reasoning. 
        Since existing geolocalization benchmarks fail to meet the need for high-resolution imagery and the localization challenge for deep agentic reasoning, we curate \textbf{GeoBench}, a benchmark that includes photos and panoramas from around the world, along with a subset of satellite images of different cities to rigorously evaluate the geolocalization ability of agentic models.
        We also propose \textbf{GeoVista}, an agentic model that seamlessly integrates tool invocation within the reasoning loop, including an image-zoom-in tool to magnify regions of interest and a web-search tool to retrieve related web information.
        We develop a complete training pipeline for it, including a cold-start supervised fine-tuning (SFT) stage to learn reasoning patterns and tool-use priors, followed by a reinforcement learning (RL) stage to further enhance reasoning ability. We adopt a hierarchical reward to leverage multi-level geographical information and improve overall geolocalization performance. 
        Experimental results show that GeoVista surpasses other open-source agentic models on the geolocalization task greatly and achieves performance comparable to closed-source models such as Gemini-2.5-flash and GPT-5 on most metrics.

        \faGithub\ \href{https://ekonwang.github.io/geo-vista/}{Webpage: ekonwang.github.io/geo-vista}
        
    \end{abstract}

    % \noindent

    \section{Introduction}

    \begin{figure*}[ht]
        \centering
        \includegraphics[width=1.\linewidth]{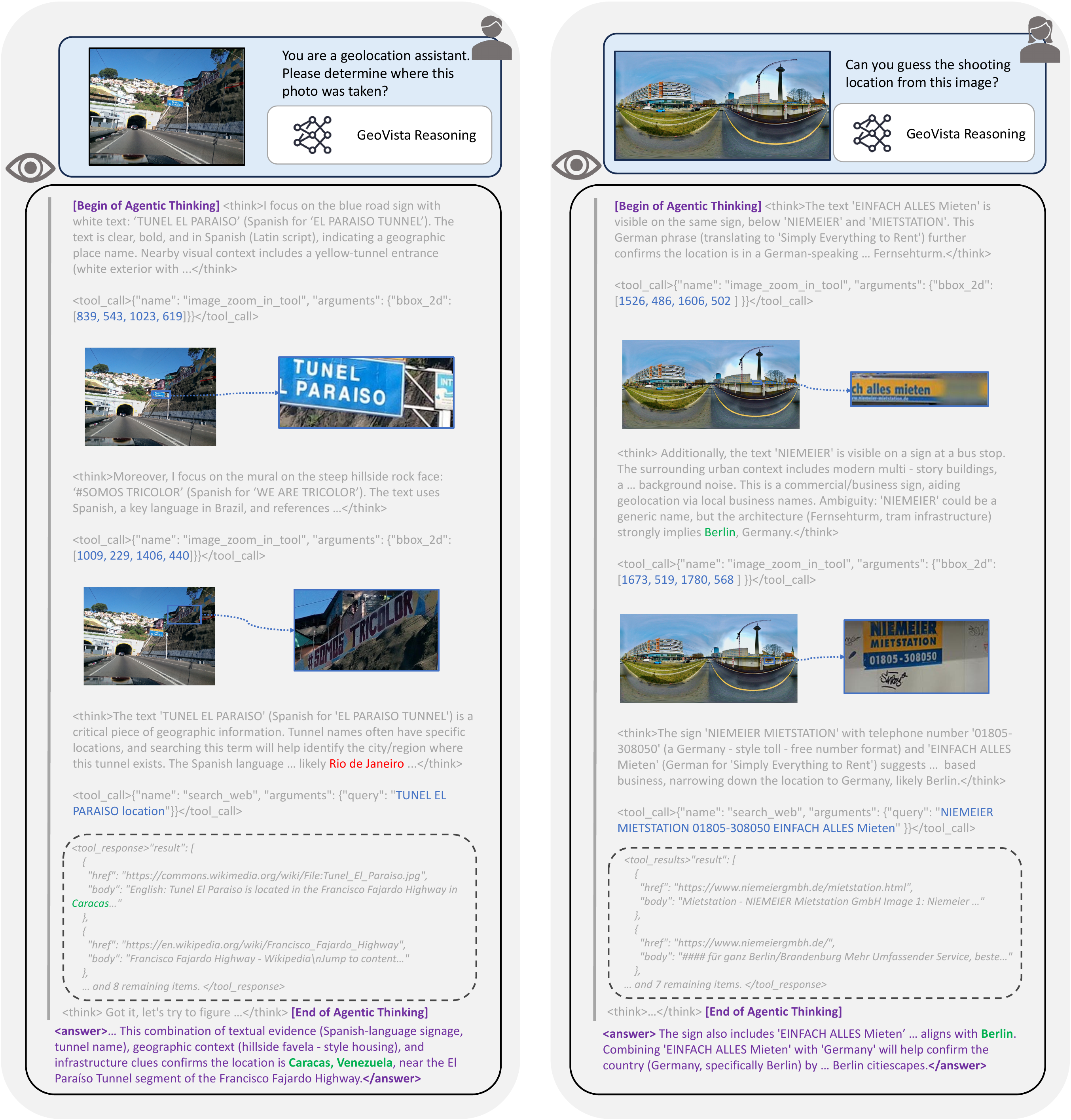}
        \caption{
        \textbf{Agentic thinking of GeoVista for Real-world geolocalization}. 
        During the reasoning loop, our GeoVista seamlessly integrates the image–zoom-in tool to magnify regions of interest and the web-search tool to retrieve relevant information. This web-augmented visual reasoning process enables GeoVista validate or refine its geolocalization judgments.
        }
        \label{fig:reasoning_trajectory}
        \vspace{5pt}
    \end{figure*}

    Recent advances in Vision-Language Models (VLMs) \citep{qwen2, deepseek_vl_v2, internvl2.5} enable deep reasoning over multimodal queries by invoking image-centric tools and utilizing long Chain-of-Thought approaches \citep{visual_cot, visual_sketchpad}, allowing these models to handle much more complex tasks. 
    Some recent works \citep{open_think_img, deepeyes} attempt to integrate seamless tool invocation into multi-turn interaction through reinforcement learning.

    Among the latest multimodal reasoning milestones~\citep{IMCoT, PeRL, simple_o3, apple_o3}, the OpenAI o3 model~\citep{OpenAI_o3_2025} enables a dynamic reasoning process with different tools integrated into it. This marks the transcendence of multimodal reasoning from one-turn queries to smooth “thinking with images” like humans, achieving a coordinated fashion of interleaving textual CoT~\citep{cot} with image manipulation and other tool invocations. Some follow-up works \citep{mini-o3, thyme} also explore combining image-centric tools with open-sourced models to achieve similar performance. However, these works only emphasize image manipulation during multimodal reasoning, thus making problem-solving rely solely on the model's inherent knowledge and lacking appropriate access to external information retrieval tools like web search.
    
    To enable a new axis for agentic multimodal reasoning, we revisit a real-world scenario—geolocalization, in which models are required to extract visual clues in high-resolution images and rely on the web search to validate or refine their hypotheses~\citep{webthinker, research, deep_research_bench}. This makes the geolocalization scenario naturally combine visual tools and information retrieval tools. To rigorously evaluate the models, we propose \textbf{GeoBench}, which consists of high-resolution photos and panoramas of global coverage. To ensure localizability as well as challenge, we remove non-localizable ones and easily recognizable landmarks. To gain insights, GeoBench also supports level-wise evaluation and nuanced evaluation to fully assess models' geolocalization capability. 
    
    We also propose our \textbf{GeoVista}, an agentic multimodal model, which seamlessly integrates tool invocation like web-search and image-zoom-in within a dynamic reasoning loop for complex geolocalization queries. As illustrated in Fig.\ref{fig:reasoning_trajectory}, GeoVista actively decides when and how to invoke tools, enabling a dynamic process of visual clue extraction and external information retrieval, reproducing reasoning behaviors similar to closed-source models like OpenAI o3. It not only utilizes visual operation and information retrieval tools to validate its hypotheses but also uses external information retrieval~\citep{open-web-research-agents, web_arena, BrowseMaster} to justify its previous wrong hypotheses and reach the correct solution. 
    % \pfliu{We propose our GeoVista, a xxx, which seamlessly xxx}
    
    We also provide a complete pipeline for GeoVista training, including cold-start and reinforcement learning. 
    First is the cold-start supervised finetuning (SFT) for learning tool-use and reasoning priors:
    We apply closed-source VLMs to generate tool invocation proposals with their rationales, execute the tool proposals to obtain the observations, and then serialize the rationales, tool invocations, and observations to generate multi-turn reasoning trajectories in order to conduct cold-start supervised finetuning. We control the number of interaction turns by limiting different tool invocation proposals.
    
    Second is the reinforcement learning to further incentivize reasoning ability~\citep{deepseek_r1}. We apply group relative policy optimization (GRPO)~\citep{grpo} with geological labels to train the models. Geological information often contains hierarchical information; to fully utilize the multi-level information, we design a hierarchical reward based on multi-level labels. This simple yet effective strategy encourages the models to learn hierarchical geological contexts from the images and make more accurate judgments.
    
    Our contributions are summarized as follows:
    
    \begin{itemize}
        \item We revisit the geolocalization task in the era of large reasoning models, which naturally requires visual clue extraction and external knowledge retrieval. We propose the \textbf{GeoBench} benchmark, which features high-resolution images with high localizability challenge, various data types of global coverage, and allows multi-level evaluation for insightful assessment.
    
        \item We propose \textbf{GeoVista}, which seamlessly integrates tool invocation within a dynamic reasoning loop for complex geolocalization queries. We also provide a complete training pipeline consisting of reasoning trajectory curation, cold-start SFT, and reinforcement learning. We further adopt a hierarchical reward during the RL stage for utilizing hierarchical information in geological labels.
    
        \item We also conduct extensive experiments to demonstrate the effectiveness of GeoVista on GeoBench and perform analysis experiments to gain insights into our approach.
    \end{itemize}

\section{Related Work}

    \subsection{Thinking with Images}

    Research on thinking with images evolved from treating images as inputs to using visual intermediates for reasoning. Visual CoT \citep{visual_cot} introduced localized intermediate steps (e.g., boxes/regions) to guide attention; Visual Sketchpad~\citep{visual_sketchpad} provided an editable canvas to draw/crop/annotate during inference; and Visual Planning argued for chains composed purely of images, replacing text with sequences of visual states. OpenAI o3 \citep{OpenAI_o3_2025} marked a watershed by productizing tool-mediated visual reasoning inside the chain (zoom, crop, rotate), triggering open replications.
    
    After the emergence of OpenAI o3 \citep{OpenAI_o3_2025}, Thyme~\citep{thyme} extends this paradigm with a code-executing visual sandbox that emits and runs image operators; mini-o3~\citep{mini-o3} trains an agent to alternate “think–act” cycles with iterative region selection and overturn masking, scaling to deep multi-turn search; OpenThinkIMG~\citep{open_think_img} unifies detectors, OCR, and drawing tools under a standardized controller with RL-learned tool policies; and DeepEyes~\citep{deepeyes} shows purely RL-induced zoom behaviors without SFT. Collectively, these systems push beyond perception toward interactive, auditable, tool-centric visual reasoning.

    \subsection{Real-World Geolocalization}

    Prior work on real-world geolocalization spans single-image, landmark, and cross-view settings. Early global photo localization built on Im2GPS~\citep{im2gps} and curated YFCC100M subsets~\citep{revisit_im2gps}, emphasizing retrieval and metric learning. Landmark-centric recognition leveraged Google Landmarks v2~\citep{google_landmarks_v2}, improving precision where distinctive structures exist. Cross-view methods advanced with VIGOR~\citep{vigor}, stressing generalization across cities for ground-to-aerial matching. Scaling to worldwide street scenes, OpenStreetView-5M~\citep{osv-5m} enabled training and fair evaluation at unprecedented diversity and size. Complementing purely visual supervision, GeoComp~\citep{geocomp} introduced human gameplay traces and reasoning sequences, catalyzing explainable, step-wise localization beyond raw appearance cues.
    Some existing works~\citep{embodiedwebagents, doxingbench, recognition_through_reasoning} also investigate agents or agentic tool uses for geo-localization.

\section{Approach}

    \begin{figure*}[t]
        \centering
        \includegraphics[width=1.\linewidth]{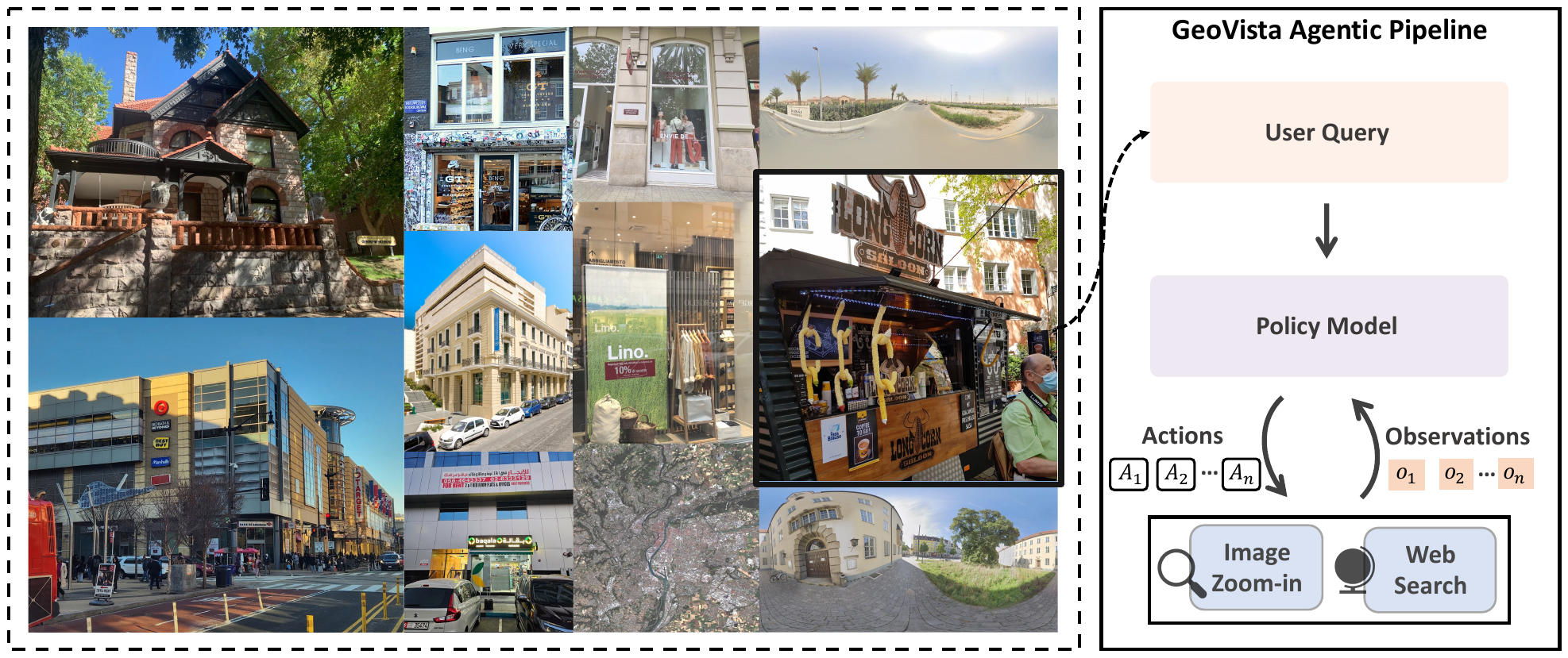}
        \caption{
        \textbf{Image examples from GeoBench and the training data, and the agentic pipeline of GeoVista.}
        Given a query and image, the policy model iteratively generates thoughts and actions; each action is parsed, executed, and yields a new observation, repeating this loop until it outputs a final geolocation prediction or reaches the maximum interaction turn limit.
        }
        \label{fig:agentic_pipeline}
        \vspace{5pt}
    \end{figure*}

    \subsection{Agentic Pipeline}

    Given a user query and an input image for geolocalization, the policy model iteratively produces a thought $T_i$ and an action $A_i$ (Fig.\ref{fig:agentic_pipeline}). The action is parsed and executed to interact with the environment, which yields a new observation $O_i$. This observation is then appended to the interaction history and fed back into the policy model. The thought–action–observation loop terminates when the model decides to present its final answer or reaches the limit of interaction turns. The tools available to the model are of two types:

    \begin{itemize}
        \item \textbf{Crop-and-Zoom.} The policy model outputs a bounding box parameterized with \texttt{bbox\_2d}, which contains pixel coordinates used to crop and magnify regions of interest. The observation is the magnified cropped subfigure.
    
        \item \textbf{Web-Search.} The policy model initializes a web search \texttt{query} to retrieve up to 10 relevant information sources from the internet. The web search service is provided by a third-party provider, and the observation consists of textual documents with web URLs.
    \end{itemize}

    \subsection{GeoBench Benchmark}

    To ensure distributional diversity, we curate \textbf{GeoBench} and training data of GeoVista from the cities worldwide. For automated labeling, each sample is accompanied by geolocalization metadata, including precise latitude and longitude. We state how we collect the raw data in Sup.\ref{sup:raw_data}.

    \paragraph{Comparison with Existing Geolocalization Benchmarks} 
    We compare our GeoBench with the existing benchmarks, we assess benchmarks along the following axes:

    \begin{table}[h]
        \centering
    
        \caption{
        \textbf{Comparison across geolocalization datasets}. 
        GeoBench is the first benchmark designed to evaluate the general geolocation ability of agentic models. It features reasonable localizability, high-resolution imagery, and hierarchical evaluation.
        }
        \label{tab:geo_dataset_comparison}
        
        \scalebox{.85}{
        \setlength{\tabcolsep}{10pt}
        \begin{tabular}{ccccccc}
        \toprule
        \textbf{Benchmark} & \textbf{Year} & \textbf{\makecell{GC}} & \textbf{\makecell{RC}} & \textbf{\makecell{HR}} & \textbf{\makecell{DV}} & \textbf{\makecell{NE}} \\
        \midrule
        \textbf{Im2GPS \citep{im2gps}}           & 2008 & \cmark &  &  &  &  \\
        \textbf{YFCC4k \citep{revisit_im2gps} }          & 2017 & \cmark &  &  &  &  \\
        \textbf{Google Landmarks v2 \citep{google_landmarks_v2}} & 2020 & \cmark &  &  &  &  \\
        \textbf{VIGOR} \citep{vigor}            & 2022 &  &  &  & \cmark & \\
        \textbf{OSV-5M} \citep{osv-5m}          & 2024 & \cmark & \cmark &  &  & \cmark  \\
        \textbf{GeoComp} \citep{geocomp}         & 2025 & \cmark & \cmark &  &  & \cmark \\
        \textbf{GeoBench} (\textit{ours})        & 2025 & \cmark & \cmark & \cmark & \cmark & \cmark \\
        \bottomrule
        \end{tabular}
        }
        \vspace{-5pt}
    \end{table}
    
    \begin{itemize}
        \item \textbf{Global Coverage.} Whether the benchmark contains images from across the globe, ensuring that the model does not overfit or bias its performance toward specific regions.
    
        \item \textbf{Reasonable Localizability.} Whether the benchmark filters out non-localizable images or easily localizable landmarks to maintain meaningful localization difficulty.
    
        \item \textbf{High Resolution.} Whether all images in the benchmark have at least $1\,\mathrm{M}$ pixels to support reliable visual clue extraction and grounding.
    
        \item \textbf{Data Variety.} Whether the benchmark includes two or more types of images to test the generalizability of reasoning models under varying data conditions.
    
        \item \textbf{Nuanced Evaluation.} Whether the benchmark includes geolocation coordinates to enable haversine distance computation for nuanced evaluation.
    \end{itemize}

    \paragraph{Localizability Filtering}
    
    We also conduct localizability filtering to remove non-localizable images and easily localizable landmarks. As we believe that images collected from the Internet exhibit varying levels of localizability~\citep{osv-5m}, especially when the data types and sources differ. Therefore, we remove two categories of data via model-based filtering:
    
    \begin{itemize}
        \item \textbf{Non-localizable images.} These images usually lack identifiable geographical clues and contain generic objects or scenes, such as close-up food photos, indoor rooms, plain natural landscapes, or single animals. Such content provides almost no regional or cultural context, making localization infeasible.
    \end{itemize}

    \begin{figure}[h!]
        \centering
        \includegraphics[width=0.7\linewidth]{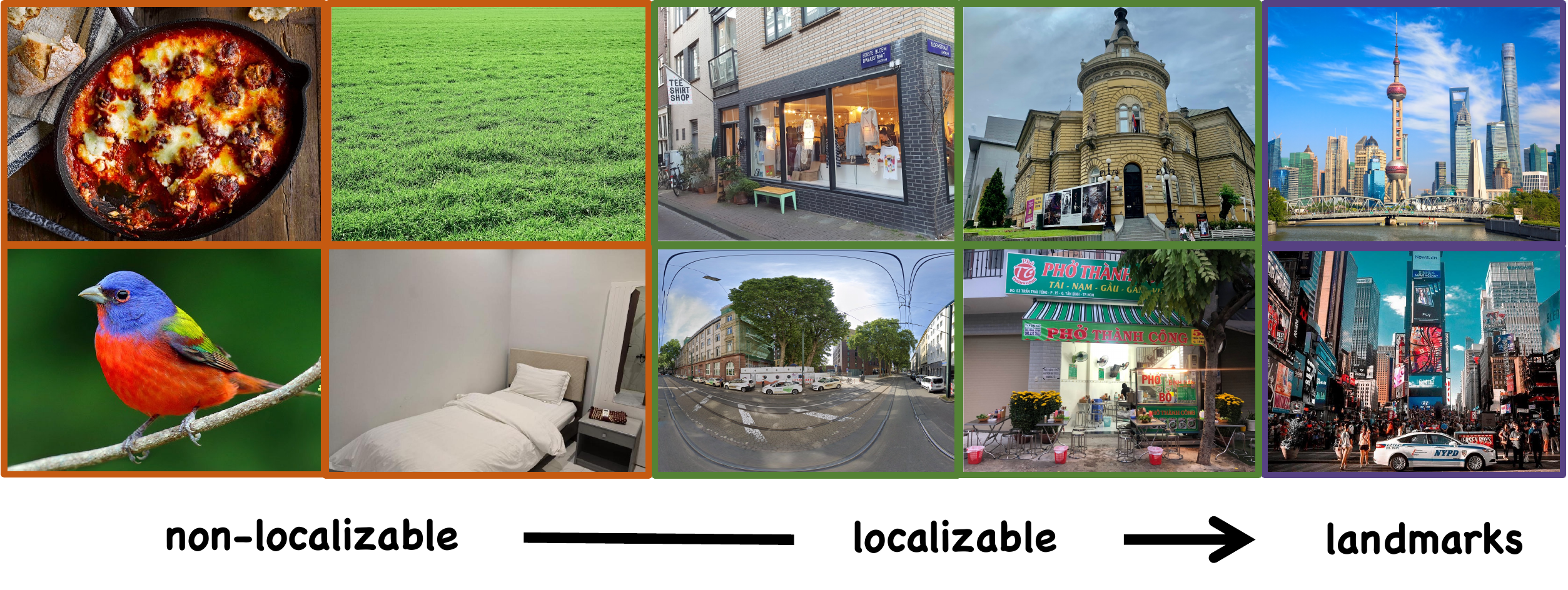}
        \vspace{-5pt}
        \caption{\textbf{Localizable vs Non-Localizable.} We remove the non-localizable (orange) and the landmarks (purple) from GeoBench, leaving only localizable images for rigorously evaluating models.}
        \label{fig:sample-efficiency}
        \vspace{-5pt}
    \end{figure}

    \begin{itemize}
        \item \textbf{Easily localizable landmarks.} These images contain strong geographic priors, typically featuring iconic landmarks or globally recognizable sites. Since VLMs have likely encountered such images multiple times during pretraining, including them would make geolocation trivial and fail to reflect genuine reasoning ability.
    \end{itemize}

    \begin{figure*}[h]
      \centering
      \begin{minipage}[t]{0.62\textwidth}
        \centering
        % \vspace{-8pt}  % 需要时微调
        \includegraphics[width=\linewidth]{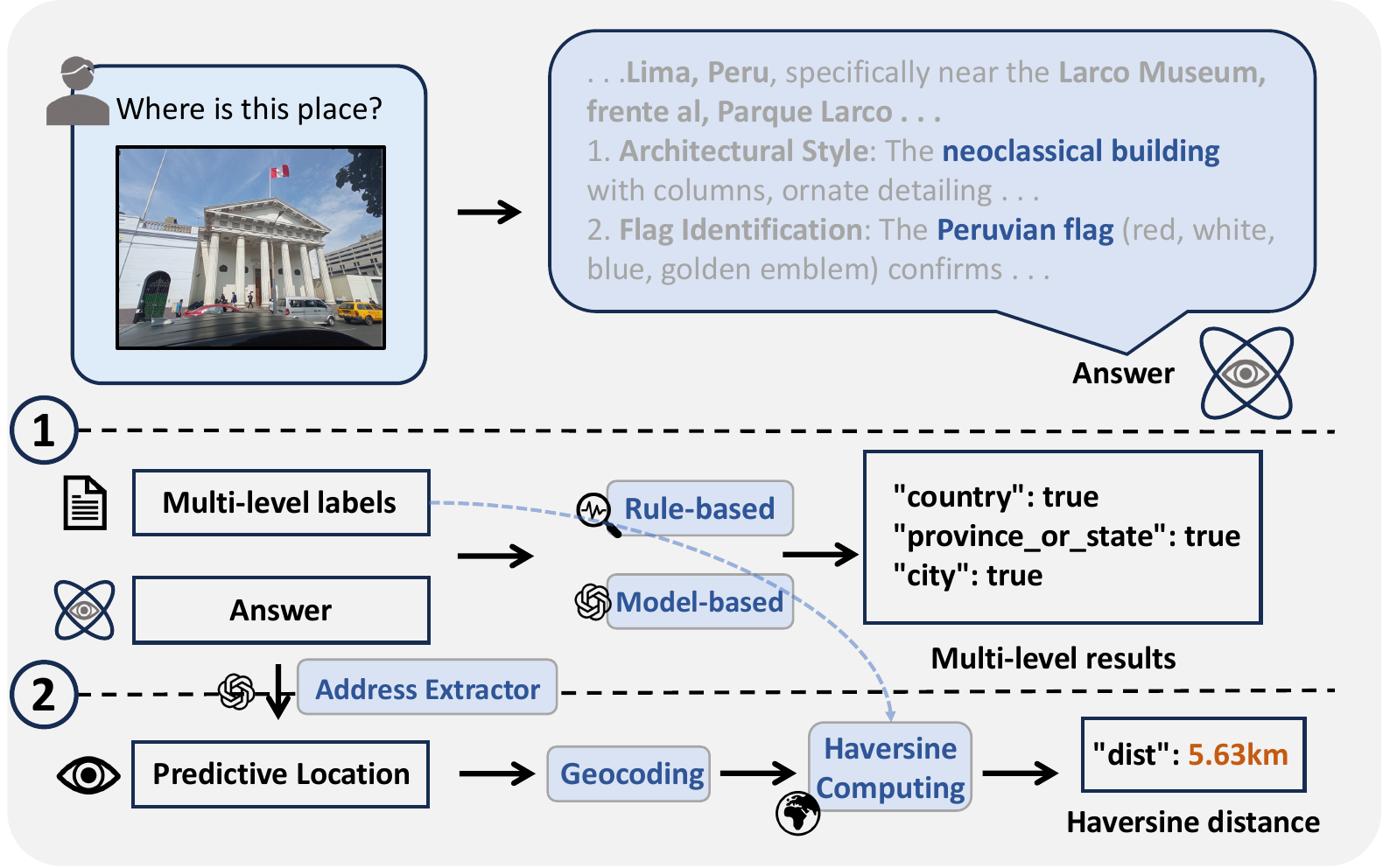}
        \label{fig:geobench-pipeline}
      \end{minipage}\hfill
      \begin{minipage}[t]{0.35\textwidth}
        \centering
        % \vspace{-8pt}
        \includegraphics[width=\linewidth]{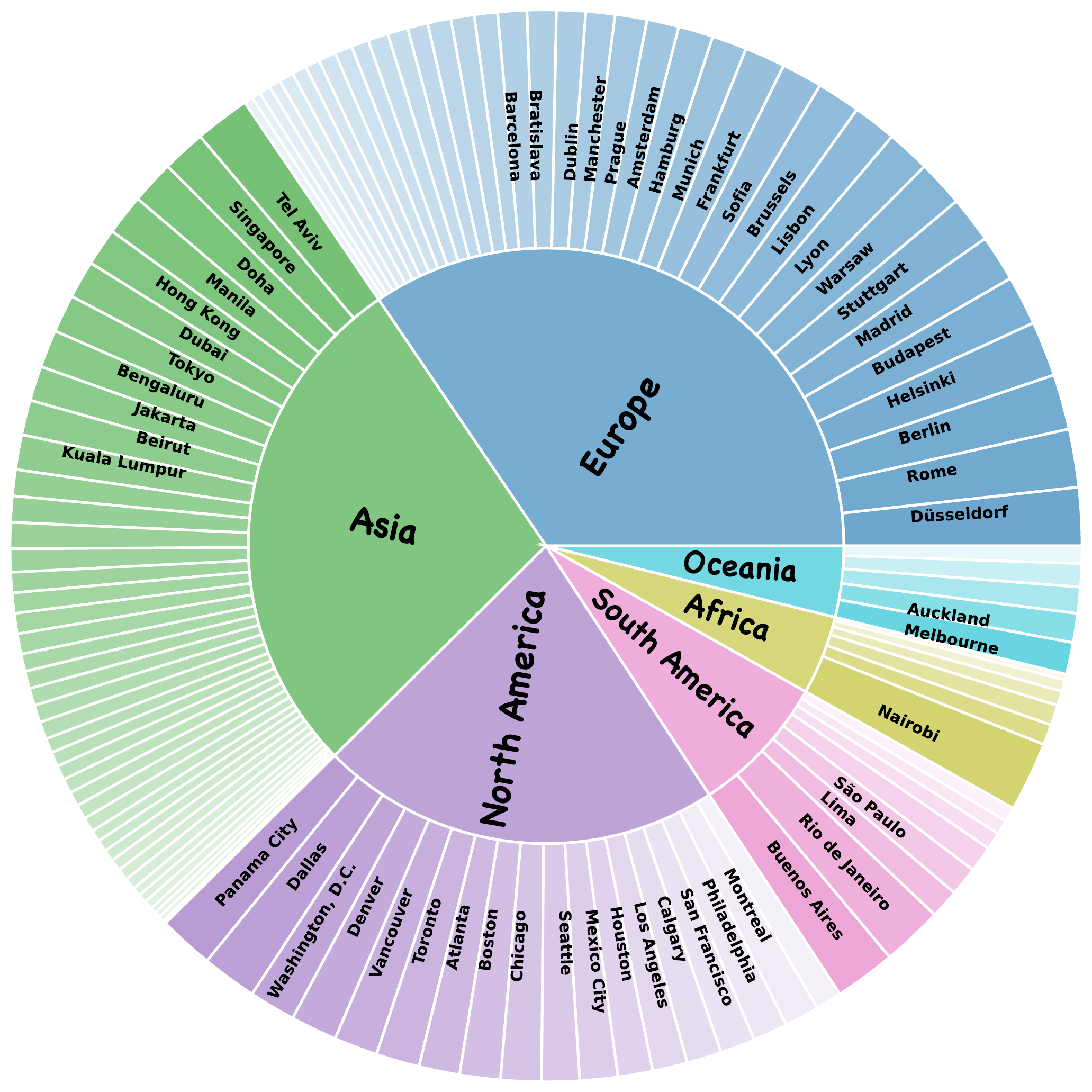}
        % \caption{\textbf{Geological distribution of GeoBench.}}
        \label{fig:geobench-distribution}
      \end{minipage}
      \caption{
      \textsc{Left}: \textbf{The evaluation pipeline of GeoBench dataset}. The evaluation system consists of (1) Level-wise evaluation, which employs both rule-based and model-based verifiers to determine correctness at different administrative levels, and (2) nuanced evaluation, which extracts the predicted address, applies geocoding to obtain the predicted geolocalization point, and computes the haversine distance to the ground-truth location. \textsc{Right}: \textbf{Geological distribution of GeoBench.} GeoBench is a high-resolution, multi-source, globally annotated dataset to evaluate models’ general geolocalization ability.
      }
      \vspace{-5pt}
      \label{fig:eval_sys_and_dist}
    \end{figure*}

    \paragraph{Level-wise Evaluation}
    To support a fully automated, rule\textendash based evaluation pipeline and to enable in\textendash depth analysis of models’ geolocalization capability, we develop multi\textendash level labels that include each image’s country, province or state, and city. With these multi\textendash level geographical labels, we combine a rule\textendash based verifier for matching specific terms with a model\textendash based verifier (using \textit{OpenAI gpt-4o-mini}) to validate the correctness of model responses at different administrative levels.

    \begin{figure*}[h]
        \centering
        \includegraphics[width=1.\linewidth]{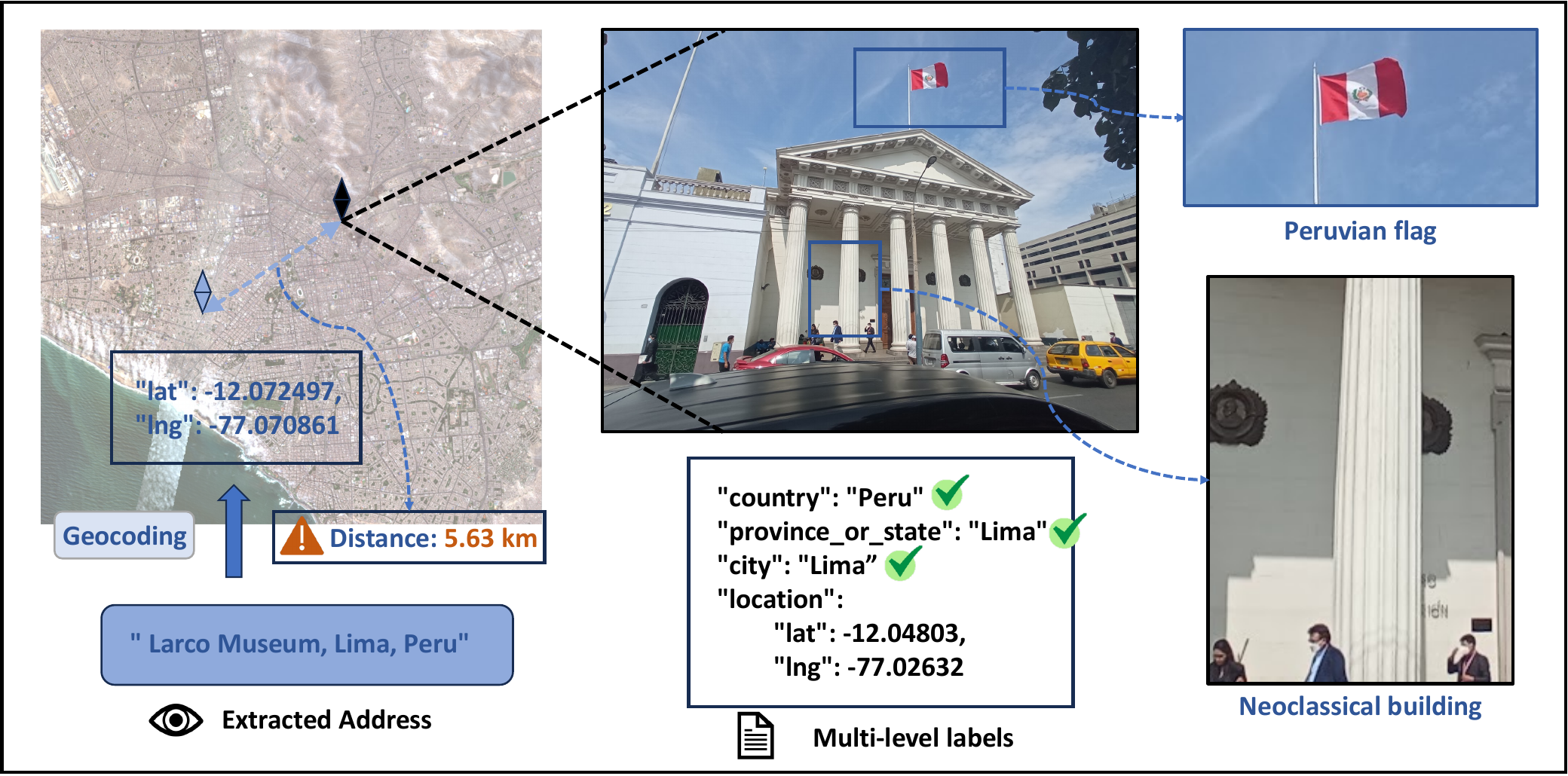}
        \caption{
        \textbf{Illustration of GeoBench dataset, along with level-wise and nuanced evaluation}. 
        }
        \label{fig:geobench}
        \vspace{-5pt}
    \end{figure*}
    
    \paragraph{Nuanced Evaluation and Haversine Distance}
    For some images with richer geographic context, state\textendash of\textendash the\textendash art (SOTA) models such as Gemini\textendash 2.5\textendash Pro can recover much more detailed addresses to street level, e.g., ``Sch\"oneberger Stra\ss e, 22149 Hamburg, Germany.'' Hence we posit that a more fine\textendash grained evaluation beyond city\textendash level is required. 
    However, models often cannot predict the geolocalization point directly, which makes nuanced evaluation difficult.
    
    To this end, as shown in Fig.\ref{fig:eval_sys_and_dist}, for each response we extract the predicted textual location and convert it into geodetic coordinates (latitude and longitude) via geocoding services (e.g., \textit{Google Geocoding API}), thereby allowing us to compute the estimated haversine distance (km) between the prediction point and the ground truth point (the geolocalization coordinates of the metadata) in an automated fasion:

    \begin{equation}
    \begin{split}
    d &= 2R_{\text{e}} \arcsin\!\left(\sqrt{v}\right), \\
    v &= \sin^2\!\left(\frac{\phi_2 - \phi_1}{2}\right)
        + \cos(\phi_1)\cos(\phi_2)
          \sin^2\!\left(\frac{\lambda_2 - \lambda_1}{2}\right)
    \end{split}
    \end{equation}
    
    where $(\phi_1, \lambda_1)$ and $(\phi_2, \lambda_2)$ are the latitude/longitude pairs of the prediction point and the ground truth point, and $R_{e}$ is Earth’s approximate radius.

    \paragraph{Geological Distribution}
    We aim to construct a dataset with diverse sources and broad geographic coverage to evaluate both closed\textendash source and open\textendash source models on general geolocalization ability. To this end, we sample $512$ standard photos, $512$ panoramas, and $108$ satellite images from the raw data (see Sup.\ref{sup:raw_data}) and conduct multi\textendash level annotation for each image. The data are high\textendash resolution to support fine\textendash grained visual reasoning, and the images span $6$ continents, $66$ countries, and $108$ cities worldwide, ranging from Xi’an to Dublin to Washington, D.C. (Fig.~\ref{fig:eval_sys_and_dist}).

    \subsection{Cold Start and Thinking Trajectory curation}

    We initially attempted to train the model (i.e., \textit{Qwen-2.5-VL-Instruct}) using reinforcement learning only, removing the need for cold-start supervised fine-tuning. However, the model tended to produce overly concise responses and hesitated to make tool calls, leading to unsatisfactory performance. This observation motivates the inclusion of explicit thinking trajectories for supervised fine-tuning, thereby incentivizing multi-turn tool-use capabilities.

\begin{figure*}[t]
        \centering
        \begin{minipage}[t]{0.48\textwidth}
            \centering
            \vspace{-0pt} % optional: adjust vertical position
            \includegraphics[width=\linewidth]{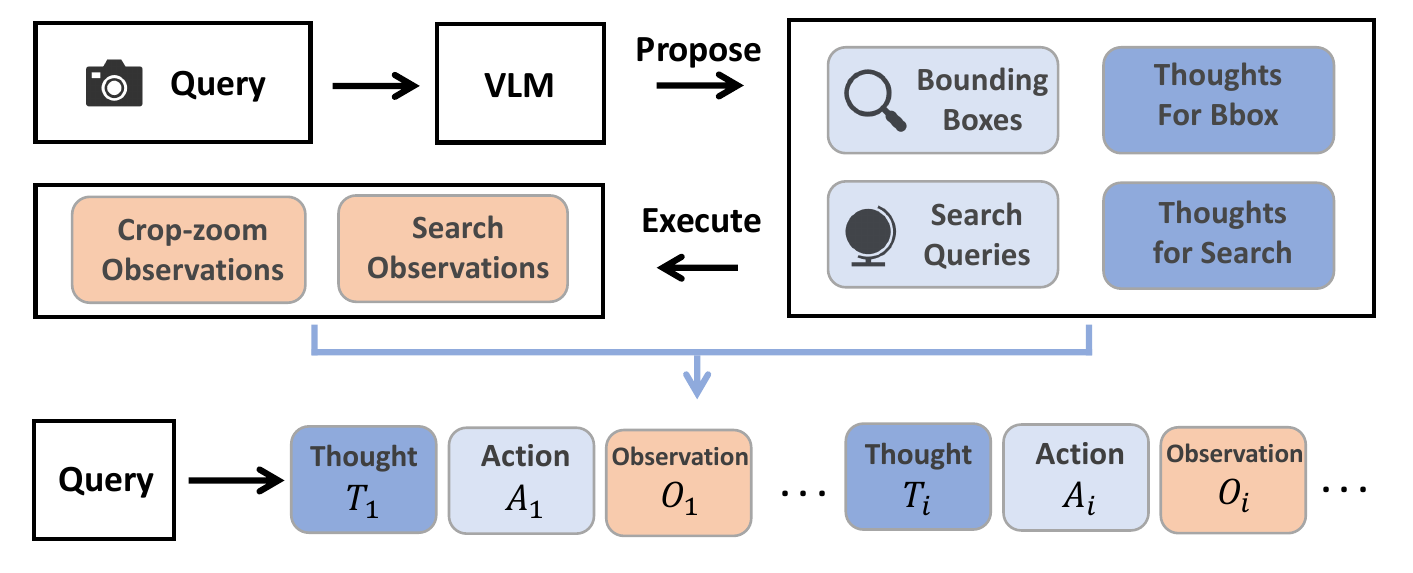}
            \label{fig:data_curation}
        \end{minipage}\hfill
        \begin{minipage}[t]{0.52\textwidth}
            \centering
            \vspace{-10pt} % optional: adjust vertical position
            \includegraphics[width=\linewidth]{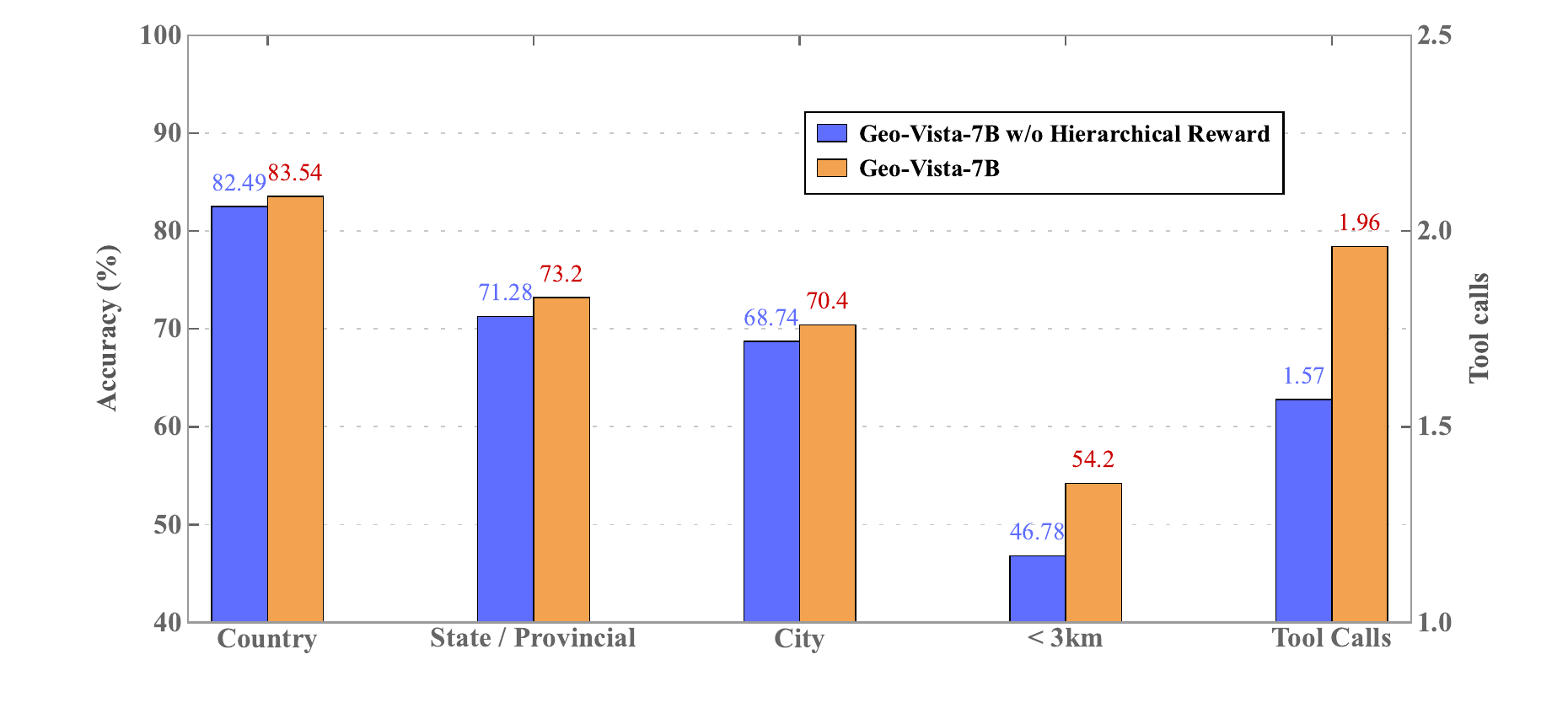}
            \label{fig:HR}
        \end{minipage}
        \vspace{-10pt}
        \caption{
        \textsc{Left}: \textbf{Thinking trajectory curation}. We mimic human geolocalization by using a VLM to propose tool calls and rationales, and assemble tool-call reasoning trajectories. \textsc{Right}: \textbf{Comparison of GeoVista-7B and its counterpart w/o Hierarchical Reward}. 
        }
        \vspace{-5pt}
        \label{fig:method_minipage}
    \end{figure*}

    Inspired by how humans identify a place during geolocalization—first selecting several candidate areas to inspect and then referencing external knowledge sources (e.g., Google Search) for further information—we inject this prior into the cold-start data. As shown in Fig.\ref{fig:method_minipage}-\textsc{Left} we use a VLM (\textit{Seed-1.6-vision}~\citep{seed_1_6_vision}) to propose multiple regions (bounding boxes) along with intermediate reasoning. After perceiving salient geographic cues, the VLM is prompted to generate several web-search queries together with the accompanying rationale, then we ask it to generate the reasoning for the final judgement.
    
    Finally, we assemble the reasoning steps, bounding boxes, and web-search queries into a coherent thinking trajectory with tool calls. As we only intend to provide the model with a reasoning pattern prior, we did not apply answer-based filtering to the reasoning trajectories. In this way, we curate $2{,}000$ cold-start reasoning trajectory examples for geolocalization.

    \subsection{Reinforcement Learning}

    We apply a vanilla GRPO~\citep{grpo} setting: each question $q$ is passed to the policy model, which generates a group of outputs $\{o_i\}_{i=1}^{G}$. Rewards $r_i$ are computed based on response correctness (e.g., whether the model predicts the city where the photo is taken). In our implementation, we do not include KL or entropy regularization. Formally, the optimization objective is:

    \begin{equation}
    \begin{aligned}
    \mathcal{J}_{\mathrm{GRPO}}(\theta)
    &= \mathbb{E}_{\,q\sim\mathcal{D},\;\{o_i\}_{i=1}^{G}\sim \pi_{\theta_{\mathrm{old}}}(\cdot\mid q)} \\
    \frac{1}{G}\sum_{i=1}^{G}&
    \Bigg[ \min\!\Bigg( 
    \frac{\pi_{\theta}(o_i\mid q)}{\pi_{\theta_{\mathrm{old}}}(o_i\mid q)}A_i,\;
    \operatorname{clip}\!\left(
    \frac{\pi_{\theta}(o_i\mid q)}{\pi_{\theta_{\mathrm{old}}}(o_i\mid q)},\,1-\epsilon,\,1+\epsilon
    \right) A_i
    \Bigg)
    \Bigg]
    \end{aligned}
    \end{equation}
    
    \begin{equation}
    \small
    A_i
    = \frac{r_i - \operatorname{mean}\!\left(\{r_1,r_2,\ldots,r_G\}\right)}
    {\operatorname{std}\!\left(\{r_1,r_2,\ldots,r_G\}\right)} \, .
    \end{equation}
    
    However, because the data have multi-level labels, a reward that only grants credit when the model predicts the correct city does not fully utilize the hierarchical information. Under this simple reward, the model underperforms on \textbf{GeoBench} and makes fewer tool calls (Fig.\ref{fig:method_minipage}-\textsc{Right}). To address this, we adopt a \textbf{hierarchical reward} to fully leverage the multi-level structure:
    
    \begin{equation}
    r_i =
        \begin{cases}
        \beta^2, & \textit{\text{if city-level correct}},\\
        \beta,   & \textit{\text{if provincial/state-level correct}},\\
        1,       & \textit{\text{if country-level correct}},\\
        0,       & \textit{else}.
        \end{cases}
    \end{equation}
    
    We set $\beta>1$ so that correct answers at smaller administrative divisions receive larger rewards. For example, for a photo taken in Los Angeles, we give a higher reward to the answer \textit{“Los Angeles”} than to \textit{“San Francisco,”} because the former is correct at the city level, although both are correct at the state level. To prevent $\beta$ from being so large that reward gaps become excessive, or so small that rewards collapse, empirically we choose a compromise value of $\beta=2$ in later experiments. 
    As reinforcement learning incurs substantial cost, particularly due to search API usage and the computational overhead of response-group rollouts, we do not experiment with additional $\beta$ values.
    
    \section{Training Recipe}
    
    \paragraph{Supervised Finetuning} During the SFT process, we use \textit{Qwen2.5-VL-7B-Instruct}~\citep{qwen2.5} as the base model. In order to avoid out-of-memory error caused by overlong trajectories, we set a max context length of 32768. We train on approximately 2000 cold-start samples for 1 epochs. The learning rate is set to $1\times10^{-5}$, with the global batch size is $32$.

    \vspace{-6pt}

    \paragraph{Reinforcement Learning} During the reinforcement learning, we employ \texttt{verl} for GRPO~\citep{grpo} implementation with $12k$ training data size. The global size is set to 64, with a mini-batch of 32. We use a constant learning rate of $1\times10^{-6}$. During the training we deprived the KL regularization~\citep{kl}. And to maintain training efficiency, we cap the maximum number of turns at 6 and set the maximum context length to 32K tokens. We also implement concurrent workers for interactions with tools during rollout to accelerate training.

    \section{Experiment}

    \subsection{Settings}

    \paragraph{Models} We compare \textbf{GeoVista} against a comprehensive suite of models. This suite is including closed-source systems—\textit{\textbf{Gemini\textendash 2.5\textendash pro}}, \textit{\textbf{Gemini\textendash 2.5\textendash flash}}~\citep{gemini-2.5-short}, \textit{\textbf{GPT-5}}~\citep{OpenAI_GPT5_2025}, and \textit{\textbf{Seed-1.6-vision}}~\citep{seed_1_6_vision}—which are supporting iterative tool calls within their reasoning process. We are also comparing open-source, vision-capable reasoning models such as \textit{\textbf{Mini-o3-7B}}~\citep{mini-o3}, \textit{\textbf{DeepEyes-7B}}~\citep{deepeyes}, and \textit{\textbf{Thyme-RL-7B}}~\citep{thyme}, which are sharing the same 7B parameter size as our \textbf{GeoVista}. We also use the base model \textit{\textbf{Qwen2.5-VL-Instruct}}~\citep{qwen2.5} for comparison. It is worth noting that the closed-source models, although not publicly specified, are likely having far larger parameter counts than 7B.

    \vspace{-6pt}

    \paragraph{Tool Use Access} We grant all open-source models identical access to the image–zoom-in tool for visual regional inspection and to a real-time web-search tool for external information retrieval. We adopt a thought–action–observation, ReAct-style~\citep{react} pattern of tool calls in multi-turn interactions. For the closed-source models like GPT-5~\citep{OpenAI_GPT5_2025}, which are already integrating comparable tools into their internal reasoning, we simply issue the query in a single turn.

    \vspace{-6pt}
    
    \paragraph{Evaluation} For a rigorous and insightful evaluation of geolocalization performance, we use \textbf{GeoBench} and conduct level-wise assessment at the \textbf{country level}, \textbf{provincial level}, and \textbf{city level}, reporting accuracy at each level. To analyze performance across different data types, we separately report city-level accuracy on panoramas, photos, and satellite images. To further assess each model’s ability to produce fine-grained geolocalization results, we conduct the nuanced evaluation and report two metrics: the proportion of predictive locations with the  \textbf{haversine distance less than $\mathbf{3\,\mathrm{\textbf{km}}}$} and the \textbf{median haversine distance}.

    \vspace{-6pt}

    \paragraph{Inference} Following the Mini-o3~\citep{mini-o3} setting, to prevent the models from being overwhelmed by the context of the original high-resolution image, we are setting the initial pixel budget to $2\,\mathrm{M}$, meaning the original image is being downsampled to at most $2\,\mathrm{M}$ pixels before entering the visual encoder.

\begin{table*}[ht]
    \centering
    \caption{
    \textbf{The Comparison on GeoBench.} The \textbf{bold} figures indicate the best performance among closed-source and open-source models, and the \underline{underlined} figures indicate open-source results that surpass at least one of their closed-source counterparts.
    }
    \label{tab:main_results}
    \scalebox{.65}{
    \setlength{\tabcolsep}{15pt}
    \begin{tabular}{l c c c c c c}
        \toprule
        \textbf{Models} & \textbf{\makecell{Country (\%) $\uparrow$} } & \textbf{\makecell{Provincial /\\State (\%) $\uparrow$}} & \textbf{\makecell{City (\%) $\uparrow$}}  & 
        \textbf{\makecell{City (\%) \\ (Panorama) $\uparrow$}} & \textbf{\makecell{City (\%) \\ (Photo) $\uparrow$}} & \textbf{\makecell{City (\%) \\ (Satellite) $\uparrow$}} \\
        \midrule
        \multicolumn{7}{c}{\textbf{Close-sourced Models}} \\
        \midrule

        \textbf{Gemini-2.5-pro} & \textbf{97.20} & \textbf{86.78} & \textbf{78.98} & \textbf{78.32} & \textbf{77.54} & \textbf{88.14} \\
        \textbf{GPT-5} & 94.09 & 77.69 & 67.11 & 69.47 & 67.92 & 53.39 \\
        \textbf{Seed-VL-1.6} & 94.31 & 81.61 & 70.58 & 69.73 & 73.44 & 61.86 \\
        % \textbf{Hunyuan-Vision-1.5}} & 91.45 & 80.24 & 70.09 & 69.1 & 71.62 & 67.02 \\
        \textbf{Gemini-2.5-flash} & 90.54 & 79.16 & 73.29 & 71.88 & 73.83 & 77.12 \\
        
        \midrule
        \multicolumn{7}{c}{\textbf{Open-sourced Models}} \\
        \midrule
        \textbf{Qwen2.5-VL-7B} & 58.93 & 42.91 & 32.57 & 24.22 & 44.73 & 16.10 \\
        \textbf{Mini-o3-7B} & 20.14  & 11.52 & 11.30 & 6.05  & 16.02 & 13.56  \\
        \textbf{DeepEyes-7B} & 54.20 & 36.08 & 30.56 & 19.92 & 42.58 & 24.58 \\
        \textbf{Thyme-RL-7B} & 69.61 & 44.31 & 30.21 & 26.17 & 35.94 & 22.88 \\
        \textbf{Geo-Vista-7B} (\textit{ours}) & \underline{\textbf{92.64}} & \underline{\textbf{79.60}} & \underline{\textbf{72.68}} & \underline{\textbf{79.49}} & \underline{\textbf{72.27}} & \textbf{44.92} \\
        
        \bottomrule
    \end{tabular}
    }
    \vspace{-5pt}
\end{table*}

    \subsection{Main Results}

    Our experimental results demonstrate GeoVista’s superior performance across metrics on GeoBench, as shown in Table~\ref{tab:main_results}. We report results at multiple geographical levels and additionally provide city-level accuracy on the GeoBench data types (i.e., panorama, photo, and satellite images). Across these metrics, GeoVista achieves state-of-the-art performance among open-source models. 
    We also find that Gemini-2.5-pro achieves the best overall performance on GeoBench among its closed-source counterparts.

    % ===== Table B: 距离相关指标（单独表） =====
% \begin{table}[ht]
%     \centering
%     \caption{
%     \textbf{Nuanced distance statistics of different models' performance on GeoBench.} The \textbf{bold} figures indicate the best performance among closed-source and open-source models.
%     }
%     \label{tab:nuanced_results}
    
%     \scalebox{.8}{
%     \setlength{\tabcolsep}{15pt}
%     \begin{tabular}{l c c}
%         \toprule
%         \textbf{Models} & \textbf{\makecell{$<$3km (\%) $\uparrow$}} & \textbf{\makecell{Median \\ Distance (km) $\downarrow$}} \\
%         \midrule
%         \multicolumn{3}{c}{\textbf{Close-sourced Models}} \\
%         \midrule

%         \textbf{Gemini-2.5-pro} & \textbf{64.45} & \textbf{0.80} \\
%         \textbf{GPT-5} & 55.12 & 1.86 \\
%         \textbf{Seed-VL-1.6} & 54.0 & 2.22 \\
%         % \textbf{Hunyuan-Vision-1.5}} & 52.49 & 2.37 \\
%         \textbf{Gemini-2.5-flash} & 58.11 & 1.67 \\
        
%         \midrule
%         \multicolumn{3}{c}{\textbf{Open-sourced Models}} \\
%         \midrule
%         \textbf{Qwen2.5-VL-7B} & 29.30 & 2209.82 \\
%         \textbf{Mini-o3-7B} & 9.57  & 13043.7 \\
%         \textbf{DeepEyes-7B} & 26.86 & 5174.93 \\
%         \textbf{Thyme-RL-7B} & 29.88 & 880.97 \\
%         \textbf{Geo-Vista-7B} (\textit{ours}) & \textbf{52.83} & \textbf{2.35} \\
        
%         \bottomrule
%     \end{tabular}
%     }
% \end{table}

\begin{wraptable}[16]{r}{0.56\textwidth}
    % \vspace{-15pt}
    \centering
    \caption{
    \textbf{Nuanced distance statistics of different models' performance on GeoBench.} 
    The \textbf{bold} figures indicate the best performance among closed-source and open-source models.
    }
    \label{tab:nuanced_results}

    \vspace{-5pt} % optional: reduce whitespace above table

    \scalebox{0.7}{
    \setlength{\tabcolsep}{15pt}
    \begin{tabular}{l c c}
        \toprule
        \textbf{Models} & \textbf{\makecell{$<$3km (\%) $\uparrow$}} & 
        \textbf{\makecell{Median \\ Distance (km) $\downarrow$}} \\
        \midrule

        \multicolumn{3}{c}{\textbf{Closed-source Models}} \\
        \midrule
        \textbf{Gemini-2.5-pro} & \textbf{64.45} & \textbf{0.80} \\
        \textbf{GPT-5} & 55.12 & 1.86 \\
        \textbf{Seed-VL-1.6} & 54.00 & 2.22 \\
        \textbf{Gemini-2.5-flash} & 58.11 & 1.67 \\

        \midrule
        \multicolumn{3}{c}{\textbf{Open-source Models}} \\
        \midrule
        \textbf{Qwen2.5-VL-7B} & 29.30 & 2209.82 \\
        \textbf{Mini-o3-7B} & 9.57  & 13043.70 \\
        \textbf{DeepEyes-7B} & 26.86 & 5174.93 \\
        \textbf{Thyme-RL-7B} & 29.88 & 880.97 \\
        \textbf{Geo-Vista-7B} (\textit{ours}) & \textbf{52.83} & \textbf{2.35} \\

        \bottomrule
    \end{tabular}
    }

    \vspace{-5pt} % optional: tighten spacing below table
\end{wraptable}

    It is worth noting that, despite having far fewer parameters, GeoVista performs on par with closed-source models on most metrics. We attribute this performance to GeoVista’s learned reasoning prior and its ability to use tool calls, especially the web-search tool. This demonstrates the effectiveness of GeoVista’s reasoning capabilities, which extend beyond simple visual grounding.

    We also conduct the \textbf{nuanced evaluation} of model predictions as shown in Tab.\ref{tab:nuanced_results}. We find that GeoVista achieves high precision for real-world geolocalization. For the two nuanced metrics we report—the rate of haversine distance $<\,3\,\mathrm{km}$ and the median haversine distance (Tab.\ref{tab:nuanced_results})—GeoVista, while leaving a small gap to closed-source models, substantially outperforms other open-source models that think with images with the same tool access, highlighting its superior reasoning performance.
    
    \subsection{Analysis}

        \subsubsection{RQ1: The Ablation Study}

        \begin{table}[ht]
    \centering

    \caption{
    \textbf{The Ablation Study.}
    Ablations on cold-start SFT, RL, and hierarchical rewards show SFT and RL are both indispensable, while hierarchical rewards further enhance multi-turn geolocalization accuracy on GeoBench.
    }
    \label{tab:ablation_results}
    
    \scalebox{0.8}{
    \setlength{\tabcolsep}{10pt}
    \begin{tabular}{l c c c c}
        \toprule
        \textbf{Models} & \textbf{\makecell{Median \\ Distance (km) $\downarrow$}} & \textbf{\makecell{City (\%) \\ (Panorama) $\uparrow$}} & \textbf{\makecell{City (\%) \\ (Photo) $\uparrow$}} & \textbf{\makecell{City (\%) \\ (Satellite) $\uparrow$}} \\
        \midrule
        \textbf{Qwen-2.5-VL} & 2209.82 & 24.22 & 44.73 & 16.1 \\
        \midrule
        \textbf{w/o Cold Start} & 55.32 & 48.52 &	43.63 &	27.46 \\
        \textbf{w/o RL}  & 11.17 & 	54.88 &	57.23 &	29.66 \\
        \textbf{w/o HR} & 4.11 &	75.0 &	68.95 &	40.68 \\
        \midrule
        \textbf{Geo-Vista-7B} & \textbf{2.35} & \textbf{79.49} & \textbf{72.27} & \textbf{44.92} \\
        
        \bottomrule
    \end{tabular}
    }

    \vspace{-5pt}
\end{table}
        
    We present an ablation study to quantify the contribution of each component. The overall results appear in Table~\ref{tab:ablation_results}. Unless otherwise stated, we keep the same training hyperparameters and evaluation settings.
    \vspace{6pt}

    \textbf{Cold Start (SFT)} \phantom{e} To assess the necessity of cold-start SFT, we remove the cold-start stage and conduct reinforcement learning directly. The results show that cold-start SFT is essential for multi-turn tool use, as performance on \textbf{GeoBench} collapses without it.
    \vspace{6pt}
    
    \textbf{Reinforcement Learning (RL)} \phantom{e} To examine the necessity of reinforcement learning, we remove the RL and only conduct the cold-start SFT. The results show that SFT alone is not sufficient: although the model learns a reasoning prior, it requires reinforcement learning to incentivize and strengthen its reasoning capability.
    \vspace{6pt}
    
    \textbf{Hierarchical Reward (HR)} \phantom{e} We also evaluate the necessity of the hierarchical reward. We keep both the cold-start SFT and reinforcement learning, but disable the hierarchical reward during the RL stage, using only a city-level reward. The results confirm the importance of hierarchical reward.
    \vspace{6pt}

        \subsubsection{RQ2: The Scaling Effect in RL Stage}

\begin{figure*}[t]
            \centering
            \begin{minipage}[t]{0.45\textwidth}
                \centering
                \vspace{-5pt} % optional: adjust vertical position
                \includegraphics[width=\linewidth]{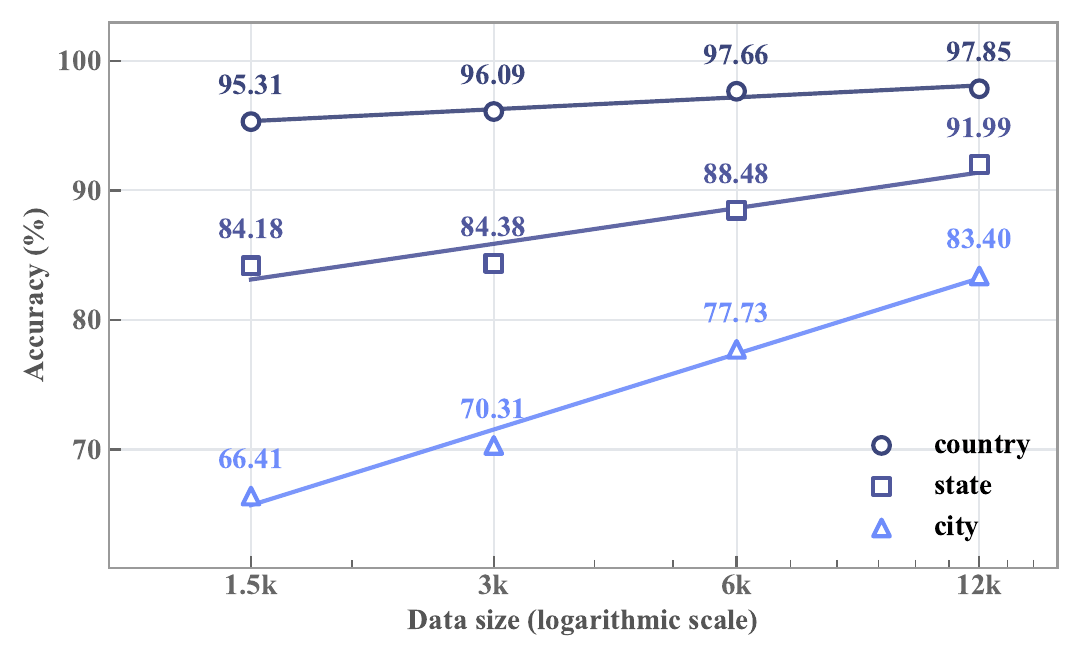}
                \label{fig:scaling_static}
            \end{minipage}\hfill
            \begin{minipage}[t]{0.55\textwidth}
                \centering
                \vspace{-10pt} % optional: adjust vertical position
                \includegraphics[width=\linewidth]{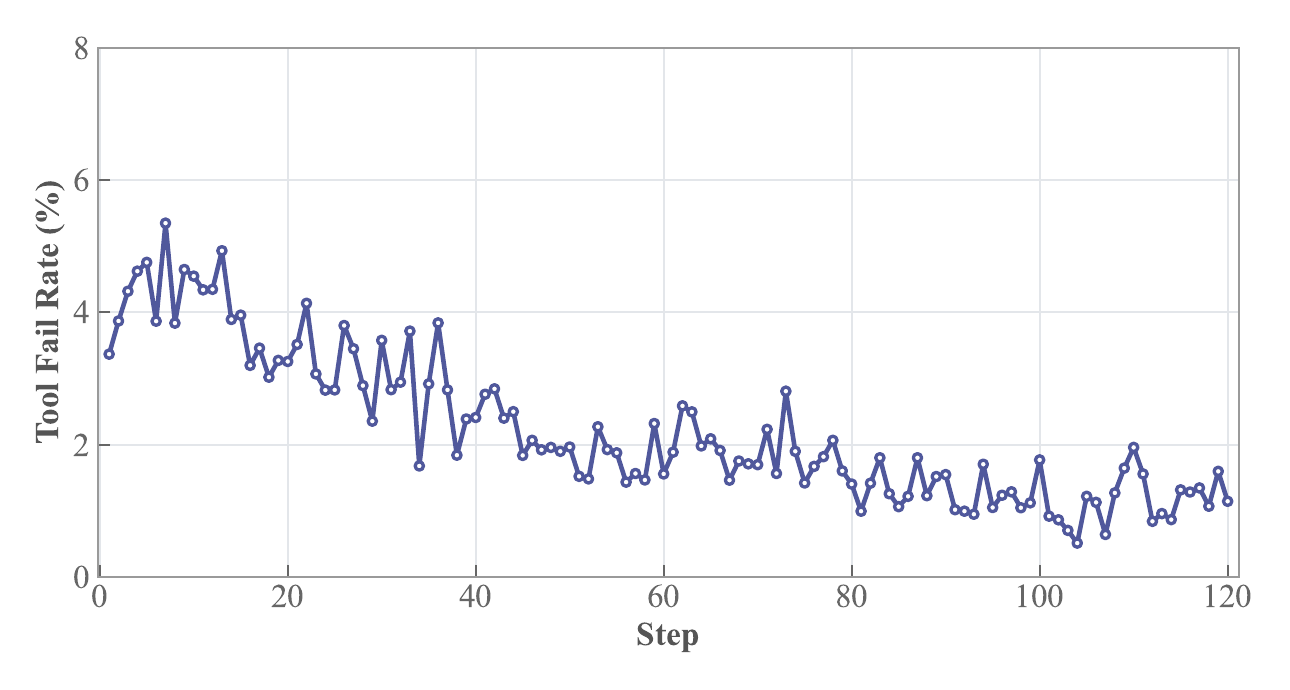}
                \label{fig:tool_fail}
            \end{minipage}
            \vspace{-10pt}
            \caption{
            \textsc{Left}: \textbf{The performance on the panorama validation set during the RL stage.} We observe nearly log-linear performance gains on the 512-panorama validation set. \textsc{Right}: \textbf{The tool fail rate during RL training.} The model’s erroneous tool-call rate decreases during RL, suggesting it learns to avoid invalid or malformed calls, leading to improved performance.
            }
            \vspace{-5pt}
            \label{fig:analysis_minipage}
        \end{figure*}

        We hypothesize that model performance increases as the data size grows. Since RL data do not require reasoning-trajectory annotations, we can easily scale the RL dataset to $12\,\mathrm{k}$ samples. We apply different RL data sizes, including $1{,}500$, $3\,\mathrm{k}$, $6\,\mathrm{k}$, and $12\,\mathrm{k}$, using the same cold-start SFT checkpoint. We report performance on a validation set consisting of $512$ panoramas. The results show that performance consistently improves as the data size increases. When plotting data size on a logarithmic scale against performance (Fig.\ref{fig:analysis_minipage}-\textsc{Left}), we observe a nearly perfect data-scaling effect.

        \subsubsection{RQ3: Failure Tool Calls during RL}
        
        To further analyze the model’s behavior regarding tool calls during RL training, we record the error tool-call rate. Error tool calls typically arise from invalid crop-tool bounding-box parameters (e.g., \texttt{x\_1} greater than \texttt{x\_2} in \texttt{bbox\_2d}) or incomplete \texttt{json} format tool-calls. An interesting observation is that, although we do not directly optimize tool-call behavior during RL, the model gradually produces fewer erroneous tool calls, showing a clear decreasing trend in error rate as training progresses (Fig.\ref{fig:analysis_minipage}-\textsc{Right}). We hypothesize that erroneous tool calls reduce the model’s likelihood of reaching the correct answer within limited turns, leading the model to implicitly learn to avoid such errors in its reasoning trajectories.

\section{Conclusion}

Our research focuses on a challenging task—real-world geolocalization—which requires searching for fine-grained visual clues and integrating external knowledge. We propose \textbf{GeoVista}, an agentic model capable of visual reasoning and tool use, including crop–zoom-in and web-search tools for deep, multi-step reasoning. To rigorously evaluate and obtain comprehensive metrics for real-world geolocalization, we introduce \textbf{GeoBench}, a benchmark containing $1{,}142$ high-resolution images from diverse global locations and three distinct data types. We curate reasoning trajectories for both cold-start supervised fine-tuning and reinforcement learning to further enhance reasoning and tool-use capabilities. We also propose a hierarchical reward to provide nuanced supervision during reinforcement learning. Experimental results show that \textbf{GeoVista} outperforms other open-source baselines and achieves performance comparable to closed-source models such as \textbf{GPT-5} and \textbf{Gemini-2.5-flash} on most metrics. Furthermore, we conduct detailed analyses for deeper insights. We believe this work lays a solid foundation for future research on agentic visual reasoning and real-world geolocalization.

\clearpage

% \clearpage

% \setcounter{page}{1}
% \maketitlesupplementary

\appendix

\section{Raw Data Collection}
\label{sup:raw_data}

\begin{figure}[h!]
    \centering
    \includegraphics[width=1\linewidth]{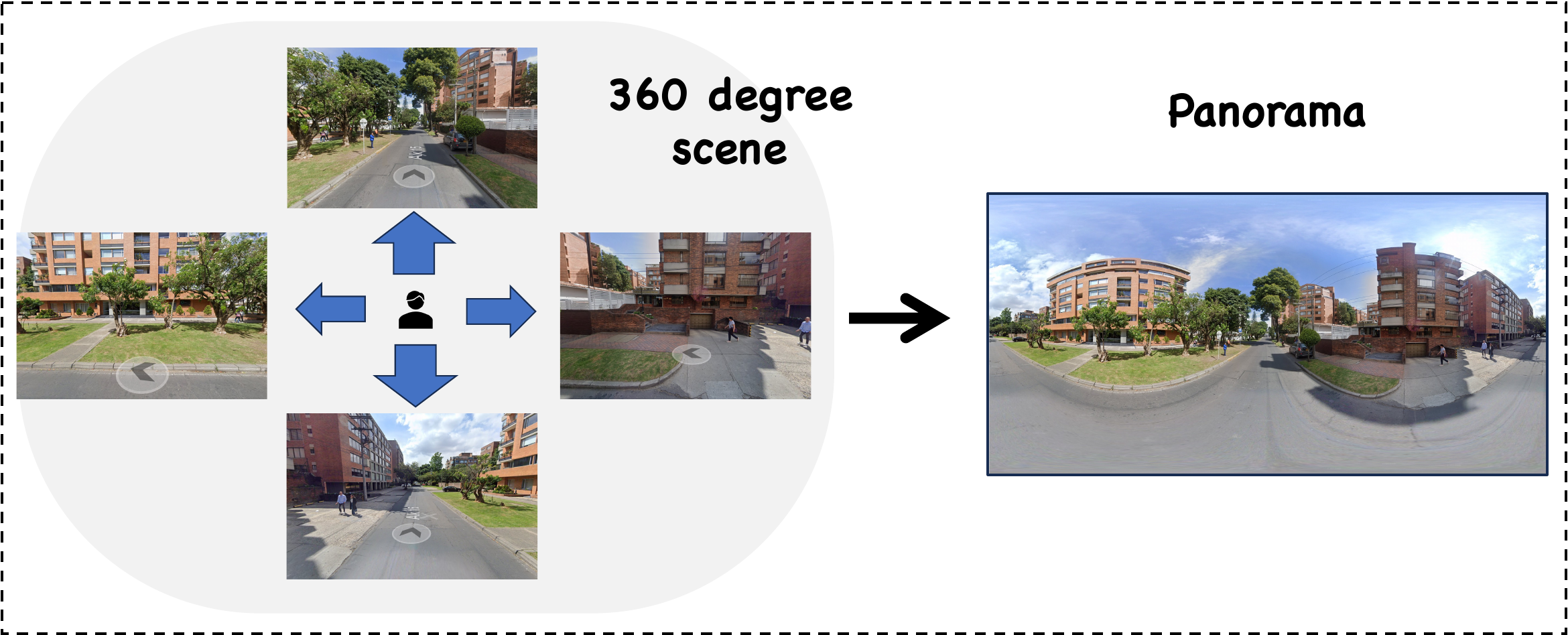}
    \caption{\textbf{The panorama pipeline in GeoBench and GeoVista training data.}}
    \label{fig:sample-efficiency}
    \vspace{-5pt}
\end{figure}

To improve the generalizability of our model rather than fitting it to a single data type, we query multiple types of raw data for GeoBench curation and training. The data types include:

    \begin{itemize}
        % \item \textbf{Normal Photos.} To obtain high-quality photographs of diverse scenarios (e.g., libraries, supermarkets), we use the Google Places API to batch-query venues and download associated images. These photos typically have resolutions of $1600\times1200$ or $1200\times1600$.

        \item \textbf{Normal Photos.} To obtain high-quality photographs of diverse scenarios (e.g., libraries, supermarkets, suburban areas), we collect photos from the internet. These photos typically have least a resolution of $1600\times1200$.
        % \item \textbf{Panoramas.} The source data are $360^\circ$ street-view scenes from cities across the globe. To make them compatible with multimodal LLM input, we convert them into planar panoramas by stitching tiles retrieved via the Google Tiles API and assembling them locally. To balance detail with storage, each panorama is fixed at a resolution of $4096\times2048$.
        \item \textbf{Panoramas.} The source data are $360^\circ$ street-view scenes from cities across the globe. To make them compatible with multimodal LLM input, we convert them into planar panoramas by stitching tiles retrieved via the Mapillary API and assembling them locally. To balance detail with storage, each panorama is fixed at a resolution of $4096\times2048$.
    \end{itemize}

    \begin{itemize}
        \item \textbf{Satellite Images.} The typical size of our satellite images is $2000\times2000$. We retrieve recent Sentinel\textendash2 Level\textendash2A imagery for cities worldwide from the Microsoft Planetary Computer, mosaic several low-cloud scenes within each city’s bounding box, and save multiple images together with their metadata from different viewport variants.
    \end{itemize}

\section{Case Study}

\begin{figure*}[h!]
    \centering
    \includegraphics[width=1.\linewidth]{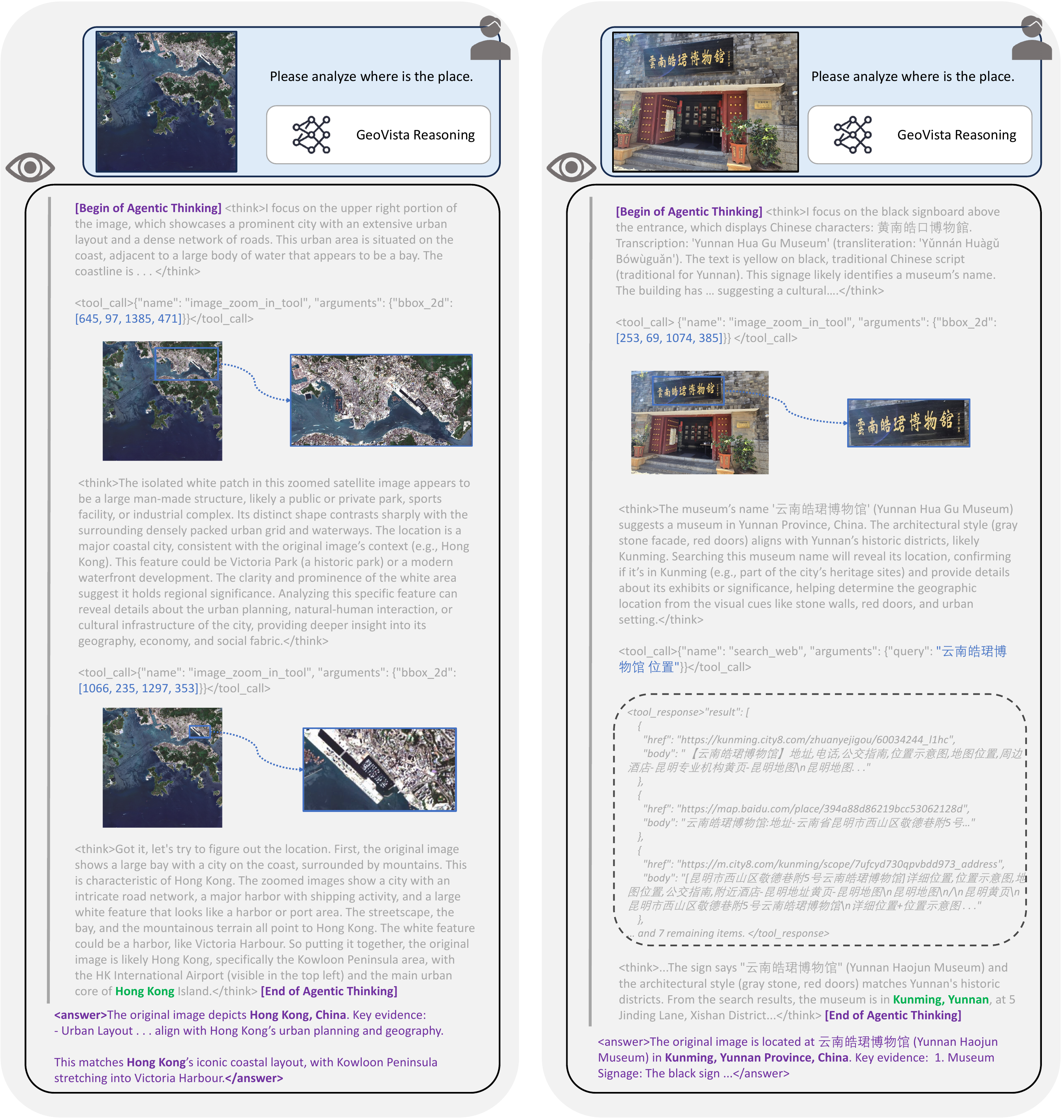}
    \caption{
    \textbf{The Reasoning Trajectory of GeoVista.} 
    We provide additional cases to facilitate the analysis of GeoVista’s reasoning trajectories and behavioral patterns, including one satellite-image example and one photo example from GeoBench.
    }
    \label{fig:case_study}
\end{figure*}

\clearpage

\section{System Prompt}

\begin{tcolorbox}[
  colback=geovistagray,
  colframe=black,
  title=\texttt{SYS PROMPT},
  fontupper=\ttfamily\small,
  sharp corners,
  boxrule=0.5pt,
  enhanced,
  breakable,
]
\begin{Verbatim}[fontsize=\scriptsize, breaklines=true]
You are a helpful assistant.
Answer the user's question based on the image provided.
# Tools

You may call one or more functions to assist with the user query.

You are provided with function signatures within <tools></tools> XML tags:

# How to call a tool
Return a json object with function name and arguments within <tool_call></tool_call> XML tags:
<tool_call>
{"name": <function-name>, "arguments": <args-json-object>}
</tool_call>

# Tool definition

## Image crop and zoom in tool
<tools>
{
"name": "image_zoom_in_tool",
"description": "Zoom in on a specific region of an image by cropping it based on a bounding box (bbox).",
"parameters": {
    "properties": {
        "bbox_2d": {
            "type": "array",
            "description": "The bounding box as [x1, y1, x2, y2] on the original image."
        }
    },
    "required": ["bbox_2d"]
}
}
</tools>

## Search web tool
<tools>
{
  "type": "function",
  "function": {
    "name": "search_web",
    "description": "Execute a web search and return normalized results containing titles, snippets, and URLs.",
    "parameters": {
      "type": "object",
      "properties": {
        "query": {
          "type": "string",
          "description": "The search query string."
        }
      },
      "required": ["query"]
    }
  }
}
</tools>


**Example**:  
<tool_call>  
{"name": "image_zoom_in_tool", "arguments": {"bbox_2d": [10, 20, 100, 200]}}  
</tool_call>
<tool_call>
{"name": "search_web", "arguments": {"query": "The palace museum"}}
</tool_call>

When you call a tool, think first. Place your internal reasoning inside <think>...</think>, followed by the <tool_call>...</tool_call>. And when you are ready to answer, also think first. Place your internal reasoning inside <think>...</think>, followed by the <answer>...</answer>.
\end{Verbatim}
\end{tcolorbox}

\clearpage

\clearpage

{
    \small
    \bibliographystyle{colm}
    \bibliography{custom}

@String(CVPR= {IEEE Conf. Comput. Vis. Pattern Recog.})

@String(ICCV= {Int. Conf. Comput. Vis.})

@String(ICLR = {Int. Conf. Learn. Represent.})

@String(CVPR  = {CVPR})

@String(ICCV  = {ICCV})

@String(ICLR  = {ICLR})

@misc{recognition_through_reasoning,
      title={Recognition through Reasoning: Reinforcing Image Geo-localization with Large Vision-Language Models}, 
      author={Ling Li and Yao Zhou and Yuxuan Liang and Fugee Tsung and Jiaheng Wei},
      year={2025},
      eprint={2506.14674},
      archivePrefix={arXiv},
      primaryClass={cs.CV},
      url={https://arxiv.org/abs/2506.14674}, 
}

@misc{doxingbench,
      title={Doxing via the Lens: Revealing Location-related Privacy Leakage on Multi-modal Large Reasoning Models}, 
      author={Weidi Luo and Tianyu Lu and Qiming Zhang and Xiaogeng Liu and Bin Hu and Yue Zhao and Jieyu Zhao and Song Gao and Patrick McDaniel and Zhen Xiang and Chaowei Xiao},
      year={2025},
      eprint={2504.19373},
      archivePrefix={arXiv},
      primaryClass={cs.CR},
      url={https://arxiv.org/abs/2504.19373}, 
}

@misc{embodiedwebagents,
      title={Embodied Web Agents: Bridging Physical-Digital Realms for Integrated Agent Intelligence}, 
      author={Yining Hong and Rui Sun and Bingxuan Li and Xingcheng Yao and Maxine Wu and Alexander Chien and Da Yin and Ying Nian Wu and Zhecan James Wang and Kai-Wei Chang},
      year={2025},
      eprint={2506.15677},
      archivePrefix={arXiv},
      primaryClass={cs.AI},
      url={https://arxiv.org/abs/2506.15677}, 
}

@misc{BrowseMaster,
      title={BrowseMaster: Towards Scalable Web Browsing via Tool-Augmented Programmatic Agent Pair}, 
      author={Xianghe Pang and Shuo Tang and Rui Ye and Yuwen Du and Yaxin Du and Siheng Chen},
      year={2025},
      eprint={2508.09129},
      archivePrefix={arXiv},
      primaryClass={cs.AI},
      url={https://arxiv.org/abs/2508.09129}, 
}

@misc{web_arena,
      title={WebArena: A Realistic Web Environment for Building Autonomous Agents}, 
      author={Shuyan Zhou and Frank F. Xu and Hao Zhu and Xuhui Zhou and Robert Lo and Abishek Sridhar and Xianyi Cheng and Tianyue Ou and Yonatan Bisk and Daniel Fried and Uri Alon and Graham Neubig},
      year={2024},
      eprint={2307.13854},
      archivePrefix={arXiv},
      primaryClass={cs.AI},
      url={https://arxiv.org/abs/2307.13854}, 
}

@misc{open-web-research-agents,
      title={Towards a Realistic Long-Term Benchmark for Open-Web Research Agents}, 
      author={Peter Mühlbacher and Nikos I. Bosse and Lawrence Phillips},
      year={2024},
      eprint={2409.14913},
      archivePrefix={arXiv},
      primaryClass={cs.CL},
      url={https://arxiv.org/abs/2409.14913}, 
}

@misc{deep_research_bench,
      title={Deep Research Bench: Evaluating AI Web Research Agents}, 
      author={FutureSearch and : and Nikos I. Bosse and Jon Evans and Robert G. Gambee and Daniel Hnyk and Peter Mühlbacher and Lawrence Phillips and Dan Schwarz and Jack Wildman},
      year={2025},
      eprint={2506.06287},
      archivePrefix={arXiv},
      primaryClass={cs.AI},
      url={https://arxiv.org/abs/2506.06287}, 
}

@misc{research,
      title={From Web Search towards Agentic Deep Research: Incentivizing Search with Reasoning Agents}, 
      author={Weizhi Zhang and Yangning Li and Yuanchen Bei and Junyu Luo and Guancheng Wan and Liangwei Yang and Chenxuan Xie and Yuyao Yang and Wei-Chieh Huang and Chunyu Miao and Henry Peng Zou and Xiao Luo and Yusheng Zhao and Yankai Chen and Chunkit Chan and Peilin Zhou and Xinyang Zhang and Chenwei Zhang and Jingbo Shang and Ming Zhang and Yangqiu Song and Irwin King and Philip S. Yu},
      year={2025},
      eprint={2506.18959},
      archivePrefix={arXiv},
      primaryClass={cs.IR},
      url={https://arxiv.org/abs/2506.18959}, 
}

@misc{webthinker,
      title={WebThinker: Empowering Large Reasoning Models with Deep Research Capability}, 
      author={Xiaoxi Li and Jiajie Jin and Guanting Dong and Hongjin Qian and Yongkang Wu and Ji-Rong Wen and Yutao Zhu and Zhicheng Dou},
      year={2025},
      eprint={2504.21776},
      archivePrefix={arXiv},
      primaryClass={cs.CL},
      url={https://arxiv.org/abs/2504.21776}, 
}

@misc{apple_o3,
      title={Interleaved Reasoning for Large Language Models via Reinforcement Learning}, 
      author={Roy Xie and David Qiu and Deepak Gopinath and Dong Lin and Yanchao Sun and Chong Wang and Saloni Potdar and Bhuwan Dhingra},
      year={2025},
      eprint={2505.19640},
      archivePrefix={arXiv},
      primaryClass={cs.CL},
      url={https://arxiv.org/abs/2505.19640}, 
}

@misc{simple_o3,
      title={Simple o3: Towards Interleaved Vision-Language Reasoning}, 
      author={Ye Wang and Qianglong Chen and Zejun Li and Siyuan Wang and Shijie Guo and Zhirui Zhang and Zhongyu Wei},
      year={2025},
      eprint={2508.12109},
      archivePrefix={arXiv},
      primaryClass={cs.CV},
      url={https://arxiv.org/abs/2508.12109}, 
}

@article{PeRL,
  author       = {Yizhen Zhang and
                  Yang Ding and
                  Shuoshuo Zhang and
                  Xinchen Zhang and
                  Haoling Li and
                  Zhong{-}zhi Li and
                  Peijie Wang and
                  Jie Wu and
                  Lei Ji and
                  Yelong Shen and
                  Yujiu Yang and
                  Yeyun Gong},
  title        = {PeRL: Permutation-Enhanced Reinforcement Learning for Interleaved
                  Vision-Language Reasoning},
  journal      = {CoRR},
  volume       = {abs/2506.14907},
  year         = {2025},
  url          = {https://doi.org/10.48550/arXiv.2506.14907},
  doi          = {10.48550/ARXIV.2506.14907},
  eprinttype    = {arXiv},
  eprint       = {2506.14907},
  bibsource    = {dblp computer science bibliography, https://dblp.org}
}

@inproceedings{IMCoT,
  author       = {Jun Gao and
                  Yongqi Li and
                  Ziqiang Cao and
                  Wenjie Li},
  title        = {Interleaved-Modal Chain-of-Thought},
  booktitle    = {{IEEE/CVF} Conference on Computer Vision and Pattern Recognition,
                  {CVPR} 2025, Nashville, TN, USA, June 11-15, 2025},
  pages        = {19520--19529},
  publisher    = {Computer Vision Foundation / {IEEE}},
  year         = {2025},
  url          = {https://openaccess.thecvf.com/content/CVPR2025/html/Gao\_Interleaved-Modal\_Chain-of-Thought\_CVPR\_2025\_paper.html},
  doi          = {10.1109/CVPR52734.2025.01818},
  bibsource    = {dblp computer science bibliography, https://dblp.org}
}

@misc{deepseek_r1,
      title={DeepSeek-R1: Incentivizing Reasoning Capability in LLMs via Reinforcement Learning}, 
      author={DeepSeek-AI and Daya Guo and Dejian Yang and Haowei Zhang and Junxiao Song and Ruoyu Zhang and Runxin Xu and Qihao Zhu and Shirong Ma and Peiyi Wang and Xiao Bi and Xiaokang Zhang and Xingkai Yu and Yu Wu and Z. F. Wu and Zhibin Gou and Zhihong Shao and Zhuoshu Li and Ziyi Gao and Aixin Liu and Bing Xue and Bingxuan Wang and Bochao Wu and Bei Feng and Chengda Lu and Chenggang Zhao and Chengqi Deng and Chenyu Zhang and Chong Ruan and Damai Dai and Deli Chen and Dongjie Ji and Erhang Li and Fangyun Lin and Fucong Dai and Fuli Luo and Guangbo Hao and Guanting Chen and Guowei Li and H. Zhang and Han Bao and Hanwei Xu and Haocheng Wang and Honghui Ding and Huajian Xin and Huazuo Gao and Hui Qu and Hui Li and Jianzhong Guo and Jiashi Li and Jiawei Wang and Jingchang Chen and Jingyang Yuan and Junjie Qiu and Junlong Li and J. L. Cai and Jiaqi Ni and Jian Liang and Jin Chen and Kai Dong and Kai Hu and Kaige Gao and Kang Guan and Kexin Huang and Kuai Yu and Lean Wang and Lecong Zhang and Liang Zhao and Litong Wang and Liyue Zhang and Lei Xu and Leyi Xia and Mingchuan Zhang and Minghua Zhang and Minghui Tang and Meng Li and Miaojun Wang and Mingming Li and Ning Tian and Panpan Huang and Peng Zhang and Qiancheng Wang and Qinyu Chen and Qiushi Du and Ruiqi Ge and Ruisong Zhang and Ruizhe Pan and Runji Wang and R. J. Chen and R. L. Jin and Ruyi Chen and Shanghao Lu and Shangyan Zhou and Shanhuang Chen and Shengfeng Ye and Shiyu Wang and Shuiping Yu and Shunfeng Zhou and Shuting Pan and S. S. Li and Shuang Zhou and Shaoqing Wu and Shengfeng Ye and Tao Yun and Tian Pei and Tianyu Sun and T. Wang and Wangding Zeng and Wanjia Zhao and Wen Liu and Wenfeng Liang and Wenjun Gao and Wenqin Yu and Wentao Zhang and W. L. Xiao and Wei An and Xiaodong Liu and Xiaohan Wang and others},
      year={2025},
      eprint={2501.12948},
      archivePrefix={arXiv},
      primaryClass={cs.CL},
      url={https://arxiv.org/abs/2501.12948}, 
}

@misc{internvl2.5,
      title={Expanding Performance Boundaries of Open-Source Multimodal Models with Model, Data, and Test-Time Scaling}, 
      author={Zhe Chen and Weiyun Wang and Yue Cao and Yangzhou Liu and Zhangwei Gao and Erfei Cui and Jinguo Zhu and Shenglong Ye and Hao Tian and Zhaoyang Liu and Lixin Gu and Xuehui Wang and Qingyun Li and Yiming Ren and Zixuan Chen and Jiapeng Luo and Jiahao Wang and Tan Jiang and Bo Wang and Conghui He and Botian Shi and Xingcheng Zhang and Han Lv and Yi Wang and Wenqi Shao and Pei Chu and Zhongying Tu and Tong He and Zhiyong Wu and Huipeng Deng and Jiaye Ge and Kai Chen and Kaipeng Zhang and Limin Wang and Min Dou and Lewei Lu and Xizhou Zhu and Tong Lu and Dahua Lin and Yu Qiao and Jifeng Dai and Wenhai Wang},
      year={2025},
      eprint={2412.05271},
      archivePrefix={arXiv},
      primaryClass={cs.CV},
      url={https://arxiv.org/abs/2412.05271}, 
}

@misc{deepseek_vl_v2,
      title={DeepSeek-VL2: Mixture-of-Experts Vision-Language Models for Advanced Multimodal Understanding}, 
      author={Zhiyu Wu and Xiaokang Chen and Zizheng Pan and Xingchao Liu and Wen Liu and Damai Dai and Huazuo Gao and Yiyang Ma and Chengyue Wu and Bingxuan Wang and Zhenda Xie and Yu Wu and Kai Hu and Jiawei Wang and Yaofeng Sun and Yukun Li and Yishi Piao and Kang Guan and Aixin Liu and Xin Xie and Yuxiang You and Kai Dong and Xingkai Yu and Haowei Zhang and Liang Zhao and Yisong Wang and Chong Ruan},
      year={2024},
      eprint={2412.10302},
      archivePrefix={arXiv},
      primaryClass={cs.CV},
      url={https://arxiv.org/abs/2412.10302}, 
}

@misc{qwen2,
      title={Qwen2-VL: Enhancing Vision-Language Model's Perception of the World at Any Resolution}, 
      author={Peng Wang and Shuai Bai and Sinan Tan and Shijie Wang and Zhihao Fan and Jinze Bai and Keqin Chen and Xuejing Liu and Jialin Wang and Wenbin Ge and Yang Fan and Kai Dang and Mengfei Du and Xuancheng Ren and Rui Men and Dayiheng Liu and Chang Zhou and Jingren Zhou and Junyang Lin},
      year={2024},
      eprint={2409.12191},
      archivePrefix={arXiv},
      primaryClass={cs.CV},
      url={https://arxiv.org/abs/2409.12191}, 
}

@inproceedings{im2gps,
  author       = {James Hays and
                  Alexei A. Efros},
  title        = {{IM2GPS:} estimating geographic information from a single image},
  booktitle    = {2008 {IEEE} Computer Society Conference on Computer Vision and Pattern
                  Recognition {(CVPR} 2008), 24-26 June 2008, Anchorage, Alaska, {USA}},
  publisher    = {{IEEE} Computer Society},
  year         = {2008},
  url          = {https://doi.org/10.1109/CVPR.2008.4587784},
  doi          = {10.1109/CVPR.2008.4587784},
  bibsource    = {dblp computer science bibliography, https://dblp.org}
}

@article{revisit_im2gps,
  title={Revisiting IM2GPS in the Deep Learning Era},
  author={Nam N. Vo and Nathan Jacobs and James Hays},
  journal={2017 IEEE International Conference on Computer Vision (ICCV)},
  year={2017},
  pages={2640-2649},
  url={https://api.semanticscholar.org/CorpusID:7449120}
}

@article{geocomp,
  author       = {Zirui Song and
                  Jingpu Yang and
                  Yuan Huang and
                  Jonathan Tonglet and
                  Zeyu Zhang and
                  Tao Cheng and
                  Meng Fang and
                  Iryna Gurevych and
                  Xiuying Chen},
  title        = {Geolocation with Real Human Gameplay Data: {A} Large-Scale Dataset
                  and Human-Like Reasoning Framework},
  journal      = {CoRR},
  volume       = {abs/2502.13759},
  year         = {2025},
  url          = {https://doi.org/10.48550/arXiv.2502.13759},
  doi          = {10.48550/ARXIV.2502.13759},
  eprinttype    = {arXiv},
  eprint       = {2502.13759},
  bibsource    = {dblp computer science bibliography, https://dblp.org}
}

@article{osv-5m,
  title={OpenStreetView-5M: The Many Roads to Global Visual Geolocation},
  author={Guillaume Astruc and Nicolas Dufour and Ioannis Siglidis and Constantin Aronssohn and Nacim Bouia and Stephanie Fu and Romain Loiseau and Van Nguyen Nguyen and Charles Raude and Elliot Vincent and Lintao Xu and Hongyu Zhou and Loic Landrieu},
  journal={2024 IEEE/CVF Conference on Computer Vision and Pattern Recognition (CVPR)},
  year={2024},
  pages={21967-21977},
  url={https://api.semanticscholar.org/CorpusID:269448726}
}

@article{vigor,
  title={VIGOR: Cross-View Image Geo-localization beyond One-to-one Retrieval},
  author={Sijie Zhu and Taojiannan Yang and Chen Chen},
  journal={2021 IEEE/CVF Conference on Computer Vision and Pattern Recognition (CVPR)},
  year={2020},
  pages={5316-5325},
  url={https://api.semanticscholar.org/CorpusID:227151840}
}

@article{google_landmarks_v2,
  title={Google Landmarks Dataset v2 – A Large-Scale Benchmark for Instance-Level Recognition and Retrieval},
  author={Tobias Weyand and Andre F. de Ara{\'u}jo and Bingyi Cao and Jack Sim},
  journal={2020 IEEE/CVF Conference on Computer Vision and Pattern Recognition (CVPR)},
  year={2020},
  pages={2572-2581},
  url={https://api.semanticscholar.org/CorpusID:214802288}
}

@misc{visual_cot,
      title={Visual CoT: Advancing Multi-Modal Language Models with a Comprehensive Dataset and Benchmark for Chain-of-Thought Reasoning}, 
      author={Hao Shao and Shengju Qian and Han Xiao and Guanglu Song and Zhuofan Zong and Letian Wang and Yu Liu and Hongsheng Li},
      year={2024},
      eprint={2403.16999},
      archivePrefix={arXiv},
      primaryClass={cs.CV},
      url={https://arxiv.org/abs/2403.16999}, 
}

@inproceedings{visual_sketchpad,
  author       = {Yushi Hu and
                  Weijia Shi and
                  Xingyu Fu and
                  Dan Roth and
                  Mari Ostendorf and
                  Luke Zettlemoyer and
                  Noah A. Smith and
                  Ranjay Krishna},
  editor       = {Amir Globersons and
                  Lester Mackey and
                  Danielle Belgrave and
                  Angela Fan and
                  Ulrich Paquet and
                  Jakub M. Tomczak and
                  Cheng Zhang},
  title        = {Visual Sketchpad: Sketching as a Visual Chain of Thought for Multimodal
                  Language Models},
  booktitle    = {Advances in Neural Information Processing Systems 38: Annual Conference
                  on Neural Information Processing Systems 2024, NeurIPS 2024, Vancouver,
                  BC, Canada, December 10 - 15, 2024},
  year         = {2024},
  url          = {http://papers.nips.cc/paper\_files/paper/2024/hash/fb82011040977c7712409fbdb5456647-Abstract-Conference.html},
  bibsource    = {dblp computer science bibliography, https://dblp.org}
}

@misc{OpenAI_o3_2025,
  author       = {OpenAI},
  title        = {Introducing OpenAI o3 and o4-mini},
  year         = {2025},
  month        = apr,
  url          = {https://openai.com/index/introducing-o3-and-o4-mini/},
  note         = {Accessed: 2025-11-13}
}

@article{thyme,
  author       = {Yifan Zhang and
                  Xingyu Lu and
                  Shukang Yin and
                  Chaoyou Fu and
                  Wei Chen and
                  Xiao Hu and
                  Bin Wen and
                  Kaiyu Jiang and
                  Changyi Liu and
                  Tianke Zhang and
                  Haonan Fan and
                  Kaibing Chen and
                  Jiankang Chen and
                  Haojie Ding and
                  Kaiyu Tang and
                  Zhang Zhang and
                  Liang Wang and
                  Fan Yang and
                  Tingting Gao and
                  Guorui Zhou},
  title        = {Thyme: Think Beyond Images},
  journal      = {CoRR},
  volume       = {abs/2508.11630},
  year         = {2025},
  url          = {https://doi.org/10.48550/arXiv.2508.11630},
  doi          = {10.48550/ARXIV.2508.11630},
  eprinttype    = {arXiv},
  eprint       = {2508.11630},
  bibsource    = {dblp computer science bibliography, https://dblp.org}
}

@article{deepeyes,
  author       = {Ziwei Zheng and
                  Michael Yang and
                  Jack Hong and
                  Chenxiao Zhao and
                  Guohai Xu and
                  Le Yang and
                  Chao Shen and
                  Xing Yu},
  title        = {DeepEyes: Incentivizing "Thinking with Images" via Reinforcement
                  Learning},
  journal      = {CoRR},
  volume       = {abs/2505.14362},
  year         = {2025},
  url          = {https://doi.org/10.48550/arXiv.2505.14362},
  doi          = {10.48550/ARXIV.2505.14362},
  eprinttype    = {arXiv},
  eprint       = {2505.14362},
  bibsource    = {dblp computer science bibliography, https://dblp.org}
}

@article{mini-o3,
  author       = {Xin Lai and
                  Junyi Li and
                  Wei Li and
                  Tao Liu and
                  Tianjian Li and
                  Hengshuang Zhao},
  title        = {Mini-o3: Scaling Up Reasoning Patterns and Interaction Turns for Visual
                  Search},
  journal      = {CoRR},
  volume       = {abs/2509.07969},
  year         = {2025},
  url          = {https://doi.org/10.48550/arXiv.2509.07969},
  doi          = {10.48550/ARXIV.2509.07969},
  eprinttype    = {arXiv},
  eprint       = {2509.07969},
  bibsource    = {dblp computer science bibliography, https://dblp.org}
}

@article{open_think_img,
  author       = {Zhaochen Su and
                  Linjie Li and
                  Mingyang Song and
                  Yunzhuo Hao and
                  Zhengyuan Yang and
                  Jun Zhang and
                  Guanjie Chen and
                  Jiawei Gu and
                  Juntao Li and
                  Xiaoye Qu and
                  Yu Cheng},
  title        = {OpenThinkIMG: Learning to Think with Images via Visual Tool Reinforcement
                  Learning},
  journal      = {CoRR},
  volume       = {abs/2505.08617},
  year         = {2025},
  url          = {https://doi.org/10.48550/arXiv.2505.08617},
  doi          = {10.48550/ARXIV.2505.08617},
  eprinttype    = {arXiv},
  eprint       = {2505.08617},
  bibsource    = {dblp computer science bibliography, https://dblp.org}
}

@article{gemini-2.5-short,
  author       = {Gemini Team},
  title        = {Gemini 2.5: Pushing the Frontier with Advanced Reasoning, Multimodality,
                  Long Context, and Next Generation Agentic Capabilities},
  journal      = {CoRR},
  volume       = {abs/2507.06261},
  year         = {2025},
  url          = {https://doi.org/10.48550/arXiv.2507.06261},
  doi          = {10.48550/ARXIV.2507.06261},
  eprinttype    = {arXiv},
  eprint       = {2507.06261},
  bibsource    = {dblp computer science bibliography, https://dblp.org}
}

@article{grpo,
  author       = {Zhihong Shao and
                  Peiyi Wang and
                  Qihao Zhu and
                  Runxin Xu and
                  Junxiao Song and
                  Mingchuan Zhang and
                  Y. K. Li and
                  Y. Wu and
                  Daya Guo},
  title        = {DeepSeekMath: Pushing the Limits of Mathematical Reasoning in Open
                  Language Models},
  journal      = {CoRR},
  volume       = {abs/2402.03300},
  year         = {2024},
  url          = {https://doi.org/10.48550/arXiv.2402.03300},
  doi          = {10.48550/ARXIV.2402.03300},
  eprinttype    = {arXiv},
  eprint       = {2402.03300},
  bibsource    = {dblp computer science bibliography, https://dblp.org}
}

@misc{OpenAI_GPT5_2025,
  author       = {OpenAI},
  title        = {Introducing GPT-5},
  year         = {2025},
  month        = aug,
  howpublished = {OpenAI Blog},
  url          = {https://openai.com/index/introducing-gpt-5/},
  note         = {Accessed: 2025-11-13}
}

@misc{qwen2.5,
      title={Qwen2.5 Technical Report}, 
      author={Qwen and : and An Yang and Baosong Yang and Beichen Zhang and Binyuan Hui and Bo Zheng and Bowen Yu and Chengyuan Li and Dayiheng Liu and Fei Huang and Haoran Wei and Huan Lin and Jian Yang and Jianhong Tu and Jianwei Zhang and Jianxin Yang and Jiaxi Yang and Jingren Zhou and Junyang Lin and Kai Dang and Keming Lu and Keqin Bao and Kexin Yang and Le Yu and Mei Li and Mingfeng Xue and Pei Zhang and Qin Zhu and Rui Men and Runji Lin and Tianhao Li and Tianyi Tang and Tingyu Xia and Xingzhang Ren and Xuancheng Ren and Yang Fan and Yang Su and Yichang Zhang and Yu Wan and Yuqiong Liu and Zeyu Cui and Zhenru Zhang and Zihan Qiu},
      year={2025},
      eprint={2412.15115},
      archivePrefix={arXiv},
      primaryClass={cs.CL},
      url={https://arxiv.org/abs/2412.15115}, 
}

@inproceedings{react,
  author       = {Shunyu Yao and
                  Jeffrey Zhao and
                  Dian Yu and
                  Nan Du and
                  Izhak Shafran and
                  Karthik R. Narasimhan and
                  Yuan Cao},
  title        = {ReAct: Synergizing Reasoning and Acting in Language Models},
  booktitle    = {The Eleventh International Conference on Learning Representations,
                  {ICLR} 2023, Kigali, Rwanda, May 1-5, 2023},
  publisher    = {OpenReview.net},
  year         = {2023},
  url          = {https://openreview.net/forum?id=WE\_vluYUL-X},
  bibsource    = {dblp computer science bibliography, https://dblp.org}
}

@misc{seed_1_6_vision,
  author       = {ByteDance Seed},
  title        = {Seed1.6 Vision},
  year         = {2025},
  month        = jun,
  howpublished = {Official Site},
  url          = {https://seed.bytedance.com/en/seed1_6},
  note         = {Accessed: 2025-11-13}
}

@book{kl,
  author    = {Cover, Thomas M. and Thomas, Joy A.},
  title     = {Elements of Information Theory},
  edition   = {2},
  publisher = {Wiley},
  year      = {2006},
  isbn      = {978-0-471-24195-9}
}

@inproceedings{cot,
  author       = {Jason Wei and
                  Xuezhi Wang and
                  Dale Schuurmans and
                  Maarten Bosma and
                  Brian Ichter and
                  Fei Xia and
                  Ed H. Chi and
                  Quoc V. Le and
                  Denny Zhou},
  editor       = {Sanmi Koyejo and
                  S. Mohamed and
                  A. Agarwal and
                  Danielle Belgrave and
                  K. Cho and
                  A. Oh},
  title        = {Chain-of-Thought Prompting Elicits Reasoning in Large Language Models},
  booktitle    = {Advances in Neural Information Processing Systems 35: Annual Conference
                  on Neural Information Processing Systems 2022, NeurIPS 2022, New Orleans,
                  LA, USA, November 28 - December 9, 2022},
  year         = {2022},
  url          = {http://papers.nips.cc/paper\_files/paper/2022/hash/9d5609613524ecf4f15af0f7b31abca4-Abstract-Conference.html},
  bibsource    = {dblp computer science bibliography, https://dblp.org}
}
}

\end{document}